\documentclass[letterpaper, 10 pt, conference]{ieeeconf}  % Comment this line out if you need a4paper
\IEEEoverridecommandlockouts                              % This command is only needed if 
\usepackage{graphics} % for pdf, bitmapped graphics files
\usepackage{epsfig} % for postscript graphics files
\usepackage{subfigure}
\usepackage{stfloats}    % To enable figures at the bottom of page
\usepackage{float}
\usepackage{amsmath} % assumes amsmath package installed
\usepackage{amssymb}  % assumes amsmath package installed
\usepackage{color} 
\usepackage{siunitx}
\usepackage{multirow}
\usepackage{hyperref}

\newcommand{\inlineeqnum}{\refstepcounter{equation}~~\mbox{(\theequation)}}

\usepackage[printonlyused,withpage,nolist,nohyperlinks]{acronym}
\begin{acronym}

\acro{NMPC}{Nonlinear Model Predictive Control}
\acro{MPC}{Model Predictive Controller}
\acro{MAV}{Micro Aerial Vehicle}
\acro{EKF}{Extended Kalman Filter}
\acro{IMU}{Inertia Measurement Unit}
\acro{GNMS}{Gauss-Newton Multiple Shooting}
\acro{SLQ}{Sequential Linear Quadratic}
\acro{MAVs}{Micro Aerial Vehicles}
\acro{DoF}{Degree of Freedom}
\acro{CoM}{Centre of Mass}
\acrodefplural{DoF}[DoFs]{Degrees of Freedom}
\acro{QP}{Quadratic Program}
\acro{ICP}{Iterative Closest Point}

\end{acronym}

\usepackage{xifthen}
\usepackage{xparse}
\usepackage{subdepth}
\usepackage{leftidx}
\usepackage{bm}
\usepackage{amsmath}
\newcommand{\norm}[1]{\left\Vert#1\right\Vert}

\newcommand{\bbm}{\begin{bmatrix}}
\newcommand{\ebm}{\end{bmatrix}}

\DeclareMathAlphabet{\mbf}{OT1}{ptm}{b}{n}
\newcommand{\mbs}[1]{{\bm{#1}}}

\newcommand{\mbsbar}[1]{{\overline{\boldsymbol{#1}}}}
\newcommand{\mbshat}[1]{{\hat{\boldsymbol{#1}}}}
\newcommand{\mbstilde}[1]{{\tilde{\boldsymbol{#1}}}}

\newcommand{\mbsdot}[1]{{\dot {\boldsymbol{#1}}}}

\newcommand{\mbfbar}[1]{{\overline{\mbf{#1}}}}
\newcommand{\mbfhat}[1]{{\hat{\mbf{#1}}}}
\newcommand{\mbftilde}[1]{{\tilde{\mbf{#1}}}}

\newcommand{\mbfdot}[1]{{\dot{\mbf{#1}}}}

\newcommand{\cframe}[1]{{\smash{\protect\underrightarrow{\mathcal{F}}_{#1}}}}

\DeclareMathAlphabet{\mathbfit}{OML}{cmm}{b}{it}

\newcommand{\argmin}{\operatornamewithlimits{argmin}}

\newcommand{\pos}[2]{\leftidx{_{#1}}{ \mbf r}{_{#2}}} % position
\newcommand{\posbar}[2]{\leftidx{_{#1}}{\mbfbar r}{_{#2}}} % position
\newcommand{\posdot}[2]{\leftidx{_{#1}}{\mbfdot r}{_{#2}}} % position

\newcommand{\vel}[3]{\leftidx{_{#1}}{\mbf v}{\IfValueTF{#2}{_{#2#3\hspace{2pt}}}{}}} % velocity
\newcommand{\veltilde}[3]{\leftidx{_{#1}}{\mbftilde v}{\IfValueTF{#2}{_{#2#3\hspace{2pt}}}{}}} % velocity
\newcommand{\velbar}[3]{\leftidx{_{#1}}{\mbfbar v}{\IfValueTF{#2}{_{#2#3\hspace{2pt}}}{}}} % velocity
\newcommand{\velhat}[3]{\leftidx{_{#1}}{\mbfhat v}{\IfValueTF{#2}{_{#2#3\hspace{2pt}}}{}}} % velocity
\newcommand{\veldot}[3]{\leftidx{_{#1}}{\mbfdot v}{\IfValueTF{#2}{_{#2#3\hspace{2pt}}}{}}} % velocity

\newcommand{\myvec}[2]{\leftidx{_{#2}\hspace{-1pt}}{\mbf #1}{}} % vector
\newcommand{\myvecbar}[2]{\leftidx{_{#2}\hspace{-1pt}}{\mbfbar #1}{}} % vector
\newcommand{\acc}[3]{\leftidx{_{#1}}{\mbf a}{\IfValueTF{#2}{_{#2#3\hspace{2pt}}}{}}} % acceleration
\newcommand{\acctilde}[3]{\leftidx{_{#1}}{\mbftilde a}{\IfValueTF{#2}{_{#2#3\hspace{2pt}}}{}}} % acceleration
\newcommand{\accbar}[3]{\leftidx{_{#1}}{\mbfbar a}{\IfValueTF{#2}{_{#2#3\hspace{2pt}}}{}}} % acceleration

\newcommand{\rotvel}[3]{\leftidx{_{#1}}{\mbs \omega}{\IfValueTF{#2}{_{#2#3\hspace{2pt}}}{}}} % rotational velocity
\newcommand{\rotveltilde}[3]{\leftidx{_{#1}}{\mbstilde \omega}{\IfValueTF{#2}{_{#2#3\hspace{2pt}}}{}}} % rotational velocity
\newcommand{\rotvelbar}[3]{\leftidx{_{#1}}{\mbsbar \omega}{\IfValueTF{#2}{_{#2#3\hspace{2pt}}}{}}} % rotational velocity
\newcommand{\rotvelhat}[3]{\leftidx{_{#1}}{\mbshat \omega}{\IfValueTF{#2}{_{#2#3\hspace{2pt}}}{}}} % rotational velocity
\newcommand{\rotveldot}[3]{\leftidx{_{#1}}{\mbsdot \omega}{\IfValueTF{#2}{_{#2#3\hspace{2pt}}}{}}} % rotational velocity derivative

\newcommand{\C}[2]{\leftidx{}{\mbf C}{_{#1#2\hspace{2pt}}}} % rotation matrix
\newcommand{\Cbar}[2]{\leftidx{}{\mbfbar C}{_{#1#2\hspace{2pt}}}} % rotation matrix
\newcommand{\q}[2]{\leftidx{}{\mbf q}{_{#1#2\hspace{2pt}}}} % quaternion of rotation
\newcommand{\qbar}[2]{\leftidx{}{\mbfbar q}{_{#1#2\hspace{2pt}}}} % quaternion of rotation
\newcommand{\qdot}[2]{\leftidx{}{\mbfdot q}{_{#1#2\hspace{2pt}}}} % quaternion of rotation

\newcommand{\sinc}{{\mathrm{sinc}}}

\title{\LARGE \bf
Nonlinear MPC with Motor Failure Identification and Recovery for Safe and Aggressive Multicopter Flight
\thanks{This work has been supported by the EPSRC grant Aerial ABM EP/N018494/1 and Imperial College London.}
}

\author{Dimos Tzoumanikas, Qingyue Yan, and Stefan Leutenegger% <-this % stops a space
\thanks{The authors are with the Smart Robotics Lab, Department of Computing, Imperial College London, UK.
        {\tt\small \{dt214, qy916, s.leutenegger\}@ic.ac.uk}}%
        \thanks{Video link: \url{https://youtu.be/cAQeSZ3tIqY}}
}

\begin{document}

\maketitle
\thispagestyle{empty}
\pagestyle{empty}

%%%%%%%%%%%%%%%%%%%%%%%%%%%%%%%%%%%%%%%%%%%%%%%%%%%%%%%%%%%%%%%%%%%%%%%%%%%%%%%%
\begin{abstract}
Safe and precise reference tracking is a crucial characteristic of \ac{MAVs} that have to operate under the influence of external disturbances in cluttered environments. In this paper, we present a \ac{NMPC} that exploits the fully physics based non-linear dynamics of the system. We furthermore show how the moment and thrust control inputs can be transformed into feasible actuator commands. In order to guarantee safe operation despite potential loss of a motor under which we show our system keeps operating safely, we developed an \ac{EKF} based motor failure identification algorithm. We verify the effectiveness of the developed pipeline in flight experiments with and without motor failures.
\end{abstract}
%%%%%%%%%%%%%%%%%%%%%%%%%%%%%%%%%%%%%%%%%%%%%%%%%%%%%%%%%%%%%%%%%%%%%%%%%%%%%%%%
\section{Introduction}
During the past years, the use of \ac{MAVs} in applications such as environmental monitoring, aerial filming, surveillance and search and rescue has dramatically increased. However, most  commonly used control algorithms lack the necessary robustness needed for the critical applications stated above and they often struggle when aggressive maneuvers are required. To some extent, these problems can be eliminated when the  \ac{MAV} model is taken into account in the control design. Model based control approaches such as \ac{MPC} have become more popular in robotics thanks to increasing computational capabilities and improved algorithmic efficiency. The design and implementation of such algorithms on real robots has become significantly easier due to the open source availability of optimisation and control toolboxes such as \cite{Houska2011a, adrlCT:SIMPAR, mattingley2012cvxgen}. At the same time, robust performance under mechanical failures (such as motor failures), can only be achieved when the failure can be correctly identified and appropriately handled by the control design. 
In our paper we  address the problem of aggressive, precise and fault tolerant  \ac{MAV} navigation.  Our contributions are as follows:
\begin{itemize}
    \item The design of a quaternion based non-linear model predictive controller with body torques and collective thrust as the control inputs. 
    \item The design of a new control allocation algorithm that maps the desired control inputs into feasible thrust commands for each motor. We consider the general case where the motors can  generate both  positive and negative thrust. Our algorithm takes into account the different motor coefficients for normal and inverted rotation eliminating the need for symmetrical propellers. We show how this can be applied in a motor failure scenario.
    \item The design of an \ac{EKF} which monitors the health of each individual actuator (motor/propeller) which we use for fast identification of an actuator failure.
    \item Seamless integration of the failure detection scheme with \ac{NMPC} and allocation under actuator failure: in particular, our hexacopter platform maintains full controllability (position and yaw) after the loss of one actuator.
    %\item We present experimental results showing how the above can be used for (i) flying aggressive maneuvers that fully exploit the system's dynamic envelope and (ii) identification and recovery after a motor failure. 
\end{itemize}
\begin{figure}[t]
\centering
\includegraphics[width=0.48\textwidth]{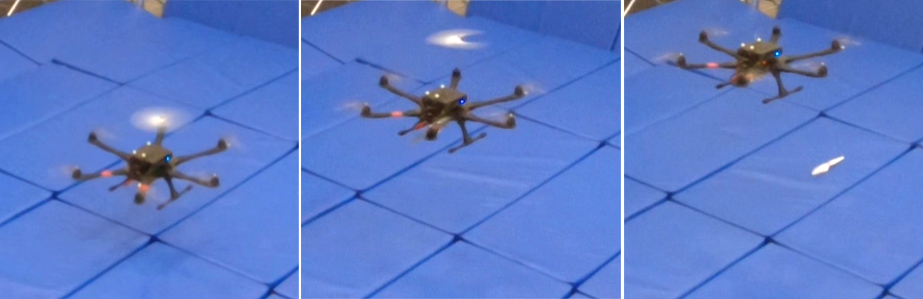}
\caption{Our  \ac{MAV} experiencing a propeller loss. The fault is identified online and the \ac{MAV} can still control its position and orientation.}
\label{fig:drone_photo}
\end{figure}
\subsection{Related Work}
Regarding \ac{MPC} in \ac{MAV}s, the most common approach is that of a cascade connection between a position and an attitude controller. In the simplest form, a linear model can be used for the translational dynamics, resulting in an optimisation problem whose solution can be solved online \cite{tzoumanikas:JFR, sa:FSR2018} or pre-computed in the form of lookup tables \cite{darivianakis:ICRA2014, papachristos:JIRS}. The use of a non-linear model for the translational dynamics such as the one presented in \cite{Kamel:IFAC2017} presents performance improvements especially when tracking of aggressive trajectories is required. The approach of the cascaded position-attitude controllers has become popular due to its ease of use, since most of the available platforms come with a pre-tuned attitude controller. However, it relies on the assumption that the attitude dynamics can be controlled independently, requiring bandwidth separation between the successive loops, i.e.\ slow control of attitude. The aforementioned works furthermore use Euler angles for the vehicle orientation which prohibits the operation close to gimbal lock.
Analogously to the position-attitude approach the authors of \cite{Falanga:IRO2018} and \cite{Foehn:ICRA2018} propose a quaternion based position controller which uses the angular rates as control inputs. These were assumed to be tracked perfectly by a separate angular rate controller.
    
The benefits of using the true non-linear model of the  \ac{MAV} has been successfully illustrated in \cite{Kamel:CCA2015} where an attitude \ac{NMPC} was employed to stabilise a hexacopter with a motor failure.
Additionally, the authors of \cite{Neunert:ICRA2016} proposed an \ac{SLQ} \ac{MPC} algorithm able to run onboard an \ac{MAV} and capable of following full state trajectory commands. Similarly, in \cite{Crousaz:ICRA2015} simulation results of an \ac{SLQ} controller stabilizing a quadrotor with slung load and a quadrotor with motor failure were presented. 

For the control allocation -- that is, the mapping of the control inputs to actuator commands -- the most commonly used method  employs the pseudo-inverse of the allocation matrix (e.g.\ \cite{Achtelik:IROS2013, Lee:CDC2010}). %The main advantage of this approach is its simplicity since the pseudo-inverse has to be computed once and the actuator commands can be obtained through a simple matrix by vector multiplication. 
In this case the actuator commands can be obtained through a simple matrix by vector multiplication. However, the main drawback is the fact that it can produce actuator commands that are not feasible. % (e.g.\ greater thrust than what each motor can provide). 
%Naively, this can be handled by clipping the actuator command such that it lies in the achievable range. 
Control allocation techniques that respect the actuator limits, such as the ones presented in \cite{Faessler:RAL2017, Brescianini:ITCST}, result in better trajectory tracking. This is partially due to the prioritisation of the roll/pitch moments and collective thrust over the yaw moment which does not directly contribute to position tracking.
Another interesting approach is the one presented in \cite{Brescianini:Mechatronics}, where the minimum energy solution is obtained by solving an optimisation problem. The authors exploit the structure of the allocation matrix nullspace in order to transform the original optimisation problem into a computationally cheaper one. Their method can be used on platforms equipped with bidirectional capable motors but requires the use of symmetrical propellers. When non symmetric propellers are used, the resulting allocation matrix is not constant but depends on the direction of rotation of each motor.

As far as fault identification and fault tolerant control are concerned, the authors of \cite{mueller:icra} were among the first to show stable position control (despite losing yaw control) with a quadrotor despite the loss of a single or two opposing propellers. The authors of \cite{schneider:fault_tolerant}  stabilised a hexacopter experiencing a motor failure. However, unlike us, they used an unconventional hexacopter motor layout which enables orientation control despite the loss of up to two motors. None of the above methods includes online fault estimation. This is done in \cite{saied:icra} where the residuals between the measured and predicted orientation and angular rate are used as criteria for detecting motor failures. Another example is the work presented in \cite{saied:ifac}, where faults are identified based on the measured motor speed and electrical current. Compared to these approaches, our method achieves up to three times faster fault detection without relying on additional sensors apart from the onboard \ac{IMU}.

\section{Notation and Definitions}
Vectors are denoted as e.g.\ $\mbf{v}$, when required with coordinate frame $\cframe{A}$ representation as $\myvec{v}{A}$. A rotation matrix $\C{A}{B}$ changes the coordinates of a vector from $\cframe{B}$ to $\cframe{A}$ as $\myvec{v}{A} = \C{A}{B}\myvec{v}{B}$. We use quaternions analogously i.e.\ $\q{A}{B}$. 
%Operator $\otimes$ denotes the quaternion multiplication.
We further denote the position of a point $P$ relative to the origin of $\cframe{A}$ as $\pos{P}{A}$. 
%We use homogeneous transformations $\T{A}{B}$ to change coordinate representations of position vectors in homogeneous coordinates as  $\posh{P}{A} = \T{A}{B}\posh{P}{B}$.
The motion of the  \ac{MAV}, with body frame $\cframe{B}$ (x: forward, y: left, z: upward), is referenced relative to the World-frame $\cframe{W}$ (z-axis upward).
\section{System Overview}
\begin{figure}[htb]
\centering
  \includegraphics[width=0.48\textwidth]{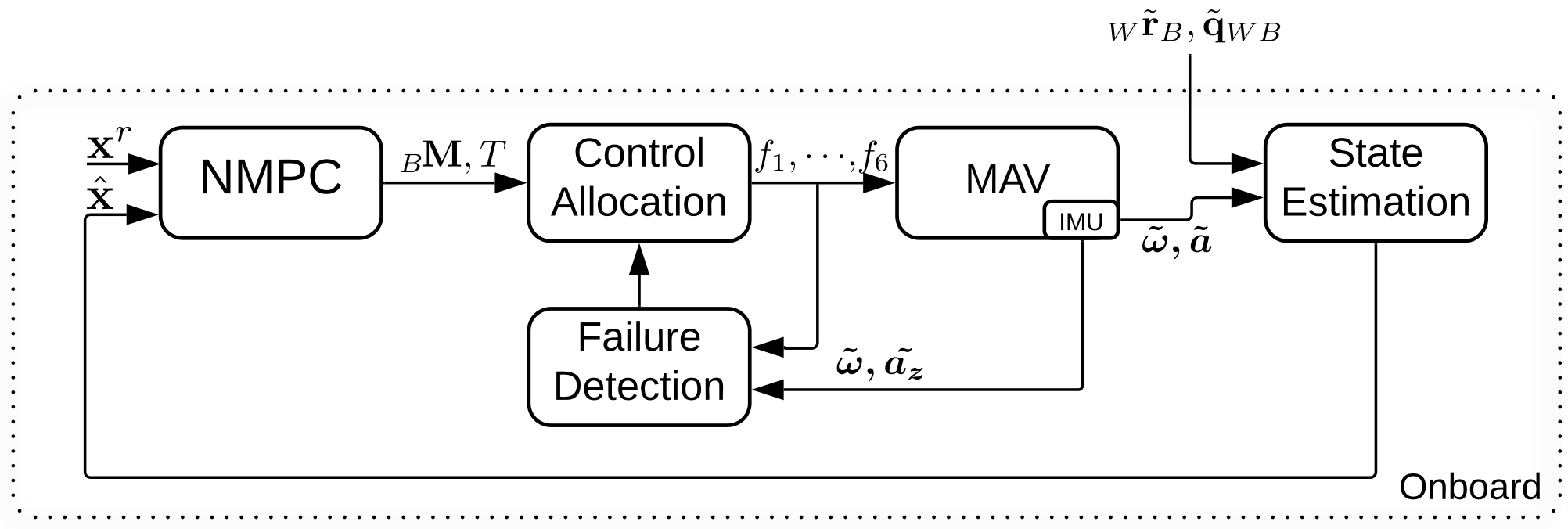}
\caption{Overview of the various software components that run onboard the \ac{MAV}. The control loop runs at 100Hz while the failure detection \ac{EKF} at 400Hz.}
\label{fig:system_overview}
\end{figure}

The software pipeline presented in this paper consists of the following different blocks: (i) the \ac{NMPC} which receives full state trajectory estimates and commands and outputs body torques and collective thrust as the control inputs; (ii) the Control allocation block which transforms the control inputs to actuator commands and finally (iii) the failure detection algorithm which estimates the health status of each motor and notifies the control allocation block in the case of a failure. An overview of the system is given in Figure \ref{fig:system_overview}.
\vspace{-1.0em}
\vspace{-1.5em}
\section{Model Based Control}
\subsection{\ac{MAV} Dynamics}
\label{sec:mav_dynamics}
\begin{figure}[htb]
\centering
  \begin{tabular}{@{}c@{}}
  \includegraphics[width=0.48\textwidth]{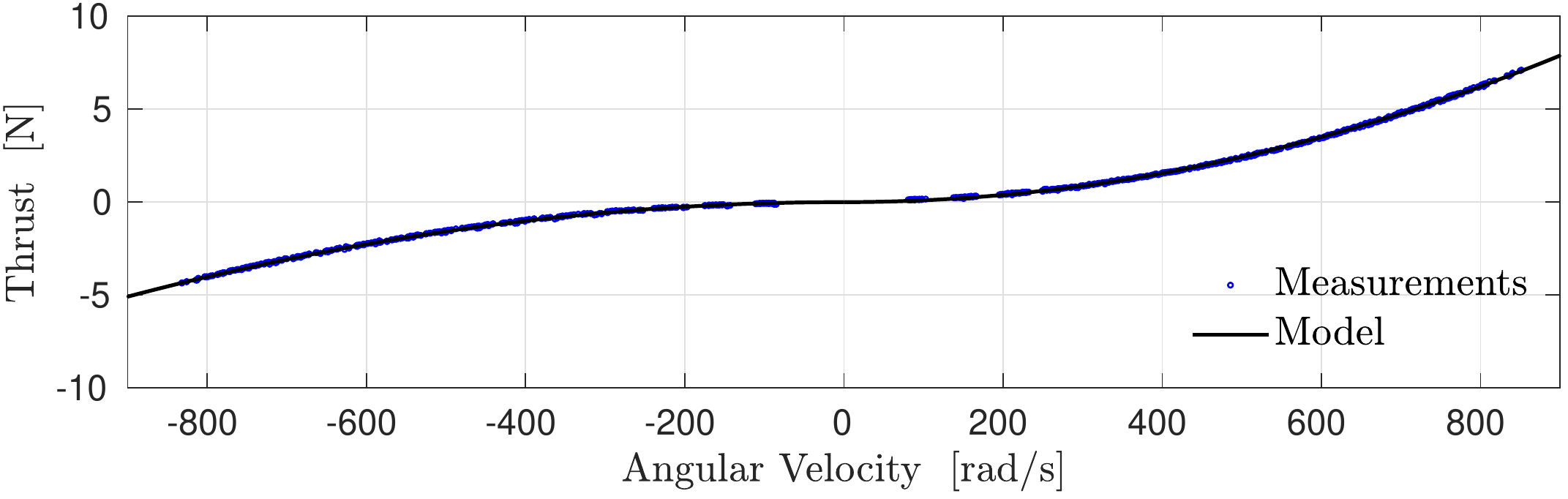} \\
  \includegraphics[width=0.48\textwidth]{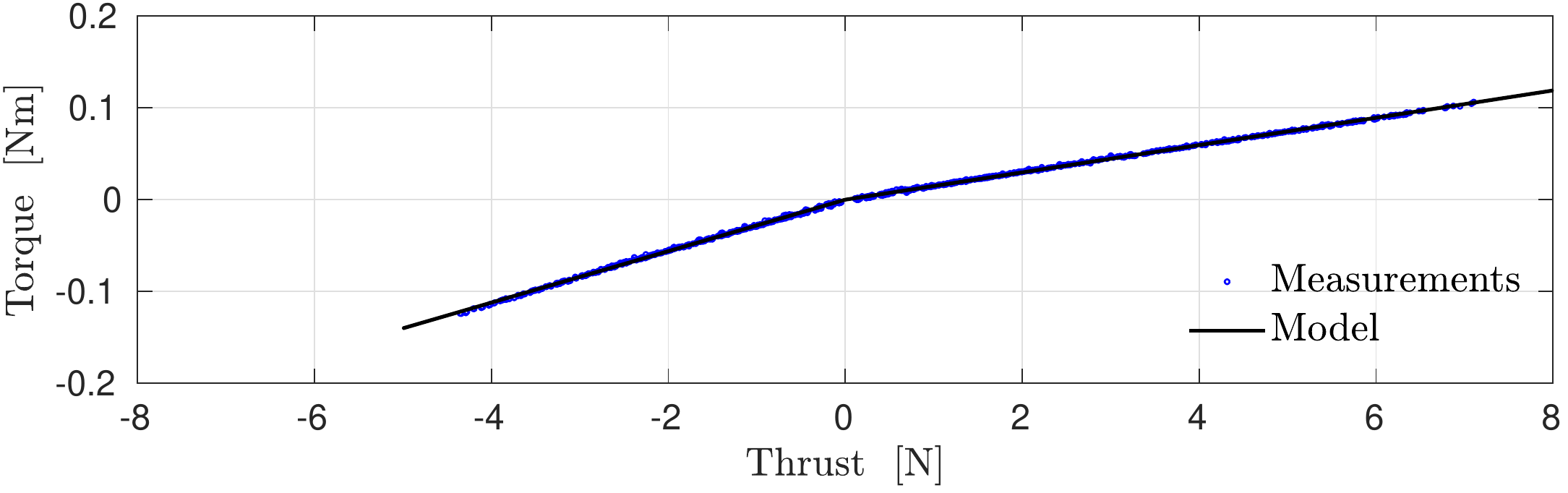}
  \end{tabular}
\caption{Experimental identification of the thrust and moment coefficients using a load cell. Each dot corresponds to the mean of 100 measurements and the solid lines correspond to the fitted model as defined in \eqref{eq:motor_model}. The use of non symmetrical propellers results in different curves for normal and inverted rotation. The least-squares fit error for the thrust and moment model for normal motor rotation is $5\times10^{-2}$ N and $8.5\times10^{-4}$ Nm respectively. The same quantities for inverted rotation are $4.4\times10^{-2}$ N and $1.3\times10^{-3}$ Nm.}
\label{fig:motor_coefficients_identification}
\end{figure}

\begin{figure}[htb]
\centering
  \includegraphics[width=0.48\textwidth]{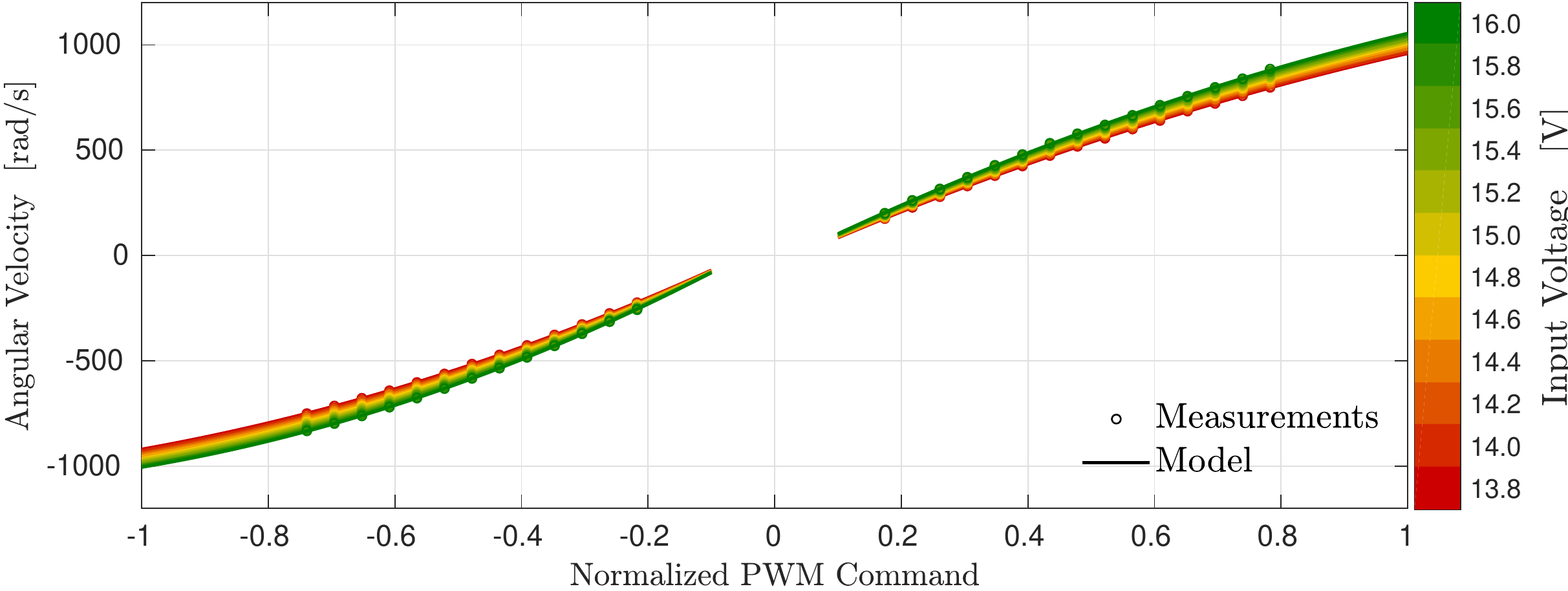}
\caption{Experimental identification of the relationship between the normalised PWM command and the achieved angular velocity. Dots correspond to averaged measurements and solid lines to the fitted quadratic polynomials. Different colours correspond to different voltage levels applied on the ESCs (here only plotting 12 different cases for visualization purposes).}
\label{fig:motor_command_identification}
\end{figure}

The Newton-Euler equations are used to model the \ac{MAV} dynamics. We ignore less significant phenomena such as the effect of the aerodynamic friction and the gyroscopic moments due to the rotation of the propellers (but our model could be extended accordingly with ease). The  \ac{MAV} dynamics can then be written in the following form: 
\begin{subequations}
\label{eq:MAV_Dynamics} 
\begin{align}
 \posdot{W}{B} &= \vel{W}{B}{}, \\ 
 \qdot{W}{B}   &= \frac{1}{2} \mathbf{\Omega}(\rotvel{B}{}{})\q{W}{B}, \\
 \veldot{W}{B}{} &= \frac{1}{m} \C{W}{B}\myvec{T}{B}+\myvec{g}{W},  \\
 \rotveldot{B}{}{} &= \mathbf{J}^{-1} (\myvec{M}{B} - \rotvel{B}{}{} \times \mathbf{J} \rotvel{B}{}{}), \\
\mathbf{\Omega} &= \begin{bmatrix}
\rotvel{B}{}{}^{\times} &  \rotvel{B}{}{} \\ 
-\rotvel{B}{}{}^{\top}   &      0
\end{bmatrix}, 
\end{align}
\end{subequations}
where $[~]^{\times}$ stands for the skew symmetric operator, $\myvec{g}{W}$ for the gravitational acceleration, $m$ for the  \ac{MAV} mass, and $\mathbf{J}$ for the inertia tensor. The thrust vector $\myvec{T}{B}:= [0, 0, T]^{\top}$ acting on the  \ac{MAV} \ac{CoM} solely depends on the collective thrust $T$ generated by the motors. This together with the moments $\myvec{M}{B}$ are considered as the control input $\myvec{u}{}:=[\myvec{M}{B},T]^{\top} \in \mathbb{R}^4$. The control state $\myvec{x}{}:=[\pos{W}{B}, \q{W}{B}, \vel{W}{B}{}, \rotvel{B}{}{}]^{\top}\in \mathbb{R}^3 \times S^3 \times \mathbb{R}^6$, consists of the \ac{MAV} position, orientation, linear and angular velocities respectively. The motor dynamics are considered significantly faster than the  \ac{MAV} body dynamics and are thus neglected. The generated thrust and moments from the $i^{\text{th}}$ motor are given by: 
\begin{subequations}
 \label{eq:motor_model}
\begin{align}
 f_i &= k_T \omega_i ^2, \\ 
 M_i &= (-1)^{i+1} k_M f_i.
\end{align}
\end{subequations}
Unlike approaches such as \cite{Faessler:RAL2017} where the relationship between the motor command and the achieved motor thrust was approximated as a quadratic polynomial, we first estimated the motor and moment coefficients defined in \eqref{eq:motor_model} and we later identified the relationship between the motor command and the achieved angular velocity. Figure \ref{fig:motor_coefficients_identification} shows the results obtained from the experimental identification of the thrust and moment coefficients $k_T$ and $k_M$ using a load cell. Since we use non symmetrical propellers which are optimised for rotation in one direction, we identified two sets of coefficients  $k_T$ and $k_M$ one for normal rotation and another one for inverted. Regarding the motor command to angular velocity relationship, we experimentally determined the dependency on input battery voltage which does not remain constant during flight.  
%We thus identified the command to angular velocity relationship for different input voltage levels. 
The identification results are illustrated in Figure \ref{fig:motor_command_identification}. The obtained quadratic polynomials for different voltage levels were stored in lookup tables and were used online depending on the measured battery voltage
\footnote{An easier and more accurate way of handling this problem is by using motors which are equipped with encoders and perform closed loop angular velocity control using the encoder information. However, the commercially available hardware, see \url{http://www.iq-control.com/} is mainly designed for small racing drones and not ones like ours which carries significant payload.}.
\vspace{-1.5em}
\subsection{Nonlinear \ac{MPC} (\ac{NMPC})}
For the control formulation, we define the following time-varying error functions for the position, linear and angular velocity, orientation and control input respectively:
\begin{subequations}
\label{eq:error_functions}
\begin{align}
    \myvec{e}{}_{r} &= \pos{W}{B} - \pos{W}{B}^{r},  \\
    \myvec{e}{}_{v} &= \vel{W}{B}{} - \vel{W}{B}{}^{r},  \\
    \myvec{e}{}_{\omega} & = \rotvel{B}{}{} -  \C{B}{B^r}\rotvel{B^r}{}{}^{r},  \\
    \myvec{e}{}_{q} &= [\q{W}{B}^{-1} \otimes \q{W}{B}^{r}]_{1:3},  \\ 
    \myvec{e}{}_{u} & = \myvec{u}{} - \myvec{u}{}^{r}. 
\end{align}
\end{subequations}
Apart from the orientation error which is obtained through quaternion multiplication, the rest corresponds to the Euclidean difference between the actual and desired (here denoted with the superscript $r$) quantity.

We compute the optimal control input $\myvec{u}{}^*$ sequence online by solving the following optimisation problem: 
\begin{subequations}
\label{eq:nloc_problem}
\begin{align}
   \myvec{u}{}^* = \argmin\limits_{\myvec{u}{}_{0}, \ldots, \myvec{u}{}_{N_f-1}} &\Big\{\Phi(\myvec{x}{}_{N_f},\myvec{x}{}^r_{N_f}) + \sum\limits_{n=0}^{N_f-1}   L(\myvec{x}{}_{n},\myvec{x}{}^r_{n},\myvec{u}{}_{n}) \Big\},  \\
    \text{s.t. : } &\myvec{x}{}_{n+1} = \myvec{f}{}_{d}(\myvec{x}{}_{n},\myvec{u}{}_{n}), \\
                   &\myvec{x}{}_{0} = \hat{\myvec{x}{}},  \\
                    &{u}_\text{lb} \leq {u}_{i} \leq {u}_\text{ub}, \quad i=1,\ldots,N, ,\end{align}
\end{subequations}
where $N_f$ is the number of time steps, $\hat{\myvec{x}{}}$ a known initial state, $\myvec{f}{}_{d}$ the discrete-time version of the  \ac{MAV} dynamics given in \eqref{eq:MAV_Dynamics} and ${u}_{\text{lb}}$, ${u}_{\text{ub}}$ lower and upper bounds for the inputs $u_{i}$. We use quadratic costs for the final and intermediate terms defined as:
\begin{subequations}
\begin{align*}
    \Phi(\myvec{x}{}_{N_f},\myvec{x}{}^r_{N_f}) = &\ \myvec{e}{}_{r}^{\top}\mathbf{Q}_{r}\myvec{e}{}_{r} + \myvec{e}{}_{v}^{\top}\mathbf{Q}_{v}\myvec{e}{}_{v} \nonumber + \myvec{e}{}_{q}^{\top}\mathbf{Q}_{q}\myvec{e}{}_{q} +  \myvec{e}{}_{\omega}^{\top}\mathbf{Q}_{\omega}\myvec{e}{}_{\omega},  \\
    L(\myvec{x}{},\myvec{x}{}^r,\myvec{u}{}) = &\ \myvec{e}{}_{r}^{\top}\mathbf{Q}_{r}\myvec{e}{}_{r} + \myvec{e}{}_{v}^{\top}\mathbf{Q}_{v}\myvec{e}{}_{v} \nonumber +
     \myvec{e}{}_{q}^{\top}\mathbf{Q}_{q}\myvec{e}{}_{q}  \\ & +  \myvec{e}{}_{\omega}^{\top}\mathbf{Q}_{\omega}\myvec{e}{}_{\omega} \nonumber +
    \myvec{e}{}_{u}^{\top}\mathbf{Q}_{u}\myvec{e}{}_{u},
\end{align*}
\end{subequations}
with $\mathbf{Q}\succcurlyeq 0$ gain matrices of appropriate dimensions which are considered tuning parameters. In our implementation we use a 10 ms discretisation step and a constant time horizon $T_f$ = 2.0 s. For the online computation of the optimal input we use the CT toolbox \cite{adrlCT:SIMPAR} and the \ac{GNMS} algorithm (outlined in \cite{Giftthaler:CoRR2017}) which result in an average computation time of 5.2 ms with standard deviation of 0.6 ms. At each GNMS iteration dynamically feasible state and input increments $\delta\myvec{x}{}$, $\delta\myvec{u}{}$ around the state and input trajectories $\bar{\myvec{X}{}}=\{\myvecbar{x}{}_{0}, \myvecbar{x}{}_{1}, \cdots, \myvecbar{x}{}_{N_f}\}$, $\bar{\myvec{U}{}}=\{\myvecbar{u}{}_{0}, \myvecbar{u}{}_{1}, \cdots, \myvecbar{u}{}_{N_f-1}\}$  are computed. 
%The \ac{GNMS} algorithm (outlined in \cite{Giftthaler:CoRR2017} is used for the solution of the nonlinear optimal control problem defined in \eqref{eq:nloc_problem}.
We obtain the dynamics of the minimal state perturbation $\mathbf{\delta\myvec{x}{}} := [\delta\pos{}{},\delta\bm{\theta}, \delta\vel{}{}{}, \delta\rotvel{}{}{}]^{\top}\in \mathbb{R}^{12}$ around the state $\bar{\myvec{x}{}}$ by introducing the local quaternion perturbation $\q{}{} = \qbar{}{} \otimes \bm{\delta}\q{}{}$ with $\bm{\delta}\q{}{} := \begin{bmatrix}
\sinc\norm{\frac{\delta\bm{\theta}}{2}}\frac{\delta\bm{\theta}}{2},\cos{\norm{\frac{\delta\bm{\theta}}{2}}}
\end{bmatrix}^\top$. The dynamics for the rotation  vector $\delta\bm{\theta} \in \mathbb{R}^3$ are given by: $\dot{\delta\bm{\theta}} = \rotvel{B}{}{} - \frac{1}{2}\rotvel{B}{}{}^{\times}\delta\bm{\theta}$, while the rotation matrix $\C{W}{B}$ can be approximated as: $\C{W}{B}\approx\Cbar{W}{B}(\mathbf{I} + \delta\bm{\theta}^{\times})$.
After each iteration the state trajectory is updated as: $\myvec{x}{} = [\posbar{w}{B } + \delta\pos{}{}, \qbar{W}{B} \otimes \delta\q{}{}, \velbar{W}{B}{}+\bm{\delta}\vel{}{}{},  \rotvelbar{B}{}{}+\delta\rotvel{}{}{}]^{\top}$.
\subsection{Control allocation}
As stated earlier, the control allocation problem involves mapping the control inputs $\myvec{u}{}^*$ to feasible actuator commands $\myvec{f}{}$. We tackle this by solving the following \ac{QP}: 
\begin{subequations}
 \label{eq:control_allocation}
\begin{align}
\myvec{f}{}^* = \argmin\limits_{\myvec{f}{}} &\Big( \left\Vert
 \mbf{A} \myvec{f}{} - \myvec{u}{}^*\right\Vert_{\mbf{W}}^{2} + \lambda\left\Vert \myvec{f}{} \right\Vert_{2}^{2}\Big) \\
 \text{s.t. : } &f_\text{min} \leq f_i \leq f_\text{max},\quad i=1,\ldots,N,
\end{align}
\end{subequations}
where $N$ is the number of motors. The allocation matrix $\mbf{A}\in\mathbb{R}^{4\times N}$, which we will present later, depends on the \ac{MAV} geometry and its motor coefficients, whereas $f_\text{min},  f_\text{max}$ correspond to the minimum and maximum attainable thrust. In order to prioritise the roll/pitch moments and the collective thrust over the yaw moment, we use the weighting matrix $\mbf{W}\in\mathbb{R}^{4\times 4}$. The scalar $\lambda\in\mathbb{R}^+$ is used such that solutions with smaller norm are preferred. When a feasible control input is commanded, solution of \eqref{eq:control_allocation} coincides with the one obtained by using the pseudo-inverse of $\mbf{A}$, namely $\myvec{f}{}=\mbf{A}^\dagger\myvec{u}{}^*$.

Since we are interested in solving the control allocation problem for the general case where the motors can produce both positive and negative thrust, we introduce the vector $\myvec{d}{}=[d_1,d_2,\cdots, d_N]$ with $d_i \in \{0, 1\}, \forall i=1,\dotsc N$. We thus use the binary variables $d_i$ to indicate whether the $i^{\text th}$ motor is spinning in its intended normal direction--corresponding to positive thrust ($d_i =  0$)--or otherwise in the inverse ($d_i =  1$).

The original optimisation problem \eqref{eq:control_allocation} is transformed to:
%into the following:
\begin{subequations}
 \label{eq:control_allocation_w_bidirectional}
\begin{align}
\myvec{f}{}^*,\myvec{d}{}^* &= \argmin\limits_{\myvec{f}{},\myvec{d}{}} \Big( \left\Vert
 {\mbf{A}(\myvec{d}{})}\myvec{f}{} - \myvec{u}{}^*\right\Vert_{\mbf{W}}^{2} + \lambda\left\Vert \myvec{f}{} \right\Vert_{2}^{2}\Big)  \\
 \text{s.t. : } &f_{\text{min}}^{+}(1-d_i) + f_{\text{min}}^{-}d_i \leq f_i \leq f_{\text{max}}^{+}(1-d_i) + f_{\text{max}}^{-}d_i,
\end{align}
\end{subequations}
where the superscript $+$ or $-$ in $f_{\text{min}}^{+}$,  $f_{\text{min}}^{-}$, $f_{\text{max}}^{+}$, $f_{\text{max}}^{-}$ has been used to indicate normal and inverted rotation respectively. The vector $\myvec{d}{}$ which encodes the direction of rotation is now an optimization variable and the allocation matrix $\mbf{A}$ is a function of $d$. For the case of the hexacopter used in our experiments, $\mbf{A}(\myvec{d}{})$ takes the following form:
%\begin{equation*}
\resizebox{.49 \textwidth}{!} {$\mbf{A}(\myvec{d}{}) = \begin{bmatrix}
                            l s_{30} & l & l s_{30} &  -l s_{30} & -l & -l s_{30} \\
                            -l c_{30} & 0 & l c_{30} &  l c_{30} & 0 & -l c_{30} \\
                            k_M(d_1) & -k_M(d_2) & k_M(d_3) &  -k_M(d_4) & k_M(d_5)  & -k_M(d_6)\\
                            1 & 1 & 1 & 1 & 1 & 1  
                           \end{bmatrix} \inlineeqnum\label{eq:allocation_matrix}$,}
where $l$ stands  for the \ac{MAV} arm length, $s_{30}=\sin{(30^o)}$, $c_{30}=\cos{(30^o)}$, $k_M(d_i) = (1-d_i)k_M^+ + d_i k_M^-$ and $k_M^+, k_M^-$ denote the normal and inverted moment coefficients identified in Section \ref{sec:mav_dynamics}.

The resulting optimisation is a mixed integer programming problem. 
However, since the possible values of $\myvec{d}{}$ are finite (72 in the case of a hexacopter), we can solve a single \ac{QP} for each single value of $\myvec{d}{}$. The global optimum $\myvec{f}{}^*$ of the optimisation problem defined in \eqref{eq:control_allocation_w_bidirectional} corresponds to the solution of the \ac{QP} with the minimum cost.
From a practical perspective solving 72 \ac{QP}s instead of a single one does not affect significantly the overall control computation time as this is dominated by the computation of the optimal input $\myvec{u}{}^*$ as described in the previous section. This is owing to the small number of optimisation variables in a single \ac{QP} tailored to the solver, CVXGEN \cite{mattingley2012cvxgen}. In our implementation, solving 72 \ac{QP}s, storing the results in a vector and finally sorting it in ascending order consistently takes less than 0.4 ms. We acknowledge, however, that our method is more resource demanding compared to methods using the pseudoinverse which can be easily implemented on a microcontroller. 

It was experimentally found that reverting the direction of rotation during flight is particularly impractical. This is because the motor dynamics are significantly slower when a direction change is commanded. As our control model does not capture this behaviour, we can prevent unnecessary direction change commands by augmenting the optimisation  \eqref{eq:control_allocation_w_bidirectional} similarly to \cite{Brescianini:Mechatronics} with the $\myvec{f}{}\in \mathcal{F}_{\text{hyst}}$ constraint, where $\mathcal{F}_{\text{hyst}}$ is the set of rotor thrusts that does not require a per motor direction change when this has already happened during the past time interval $t_{\text{hyst}}$. The solution satisfying this constraint can be found with a single iteration over the vector of 72 possibilities. The threshold $t_{\text{hyst}}$ can be iteratively decreased until a good (e.g.\ $\left\Vert\mbf{A} \myvec{f}{} - \myvec{u}{}^*\right\Vert_{\mbf{W}}^{2} < \epsilon$) solution is found.
\section{Fault Identification}
\begin{figure}[htb]
\centering
  \includegraphics[width=0.48\textwidth]{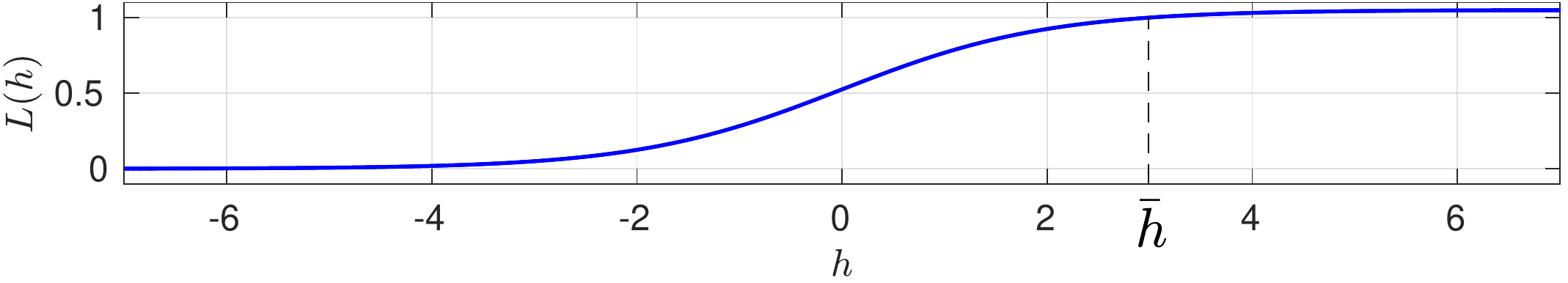} 
\caption{The logistic function used for the \ac{EKF}. Notice, that it is appropriately scaled such that $\bar{h} = L^{-1}(1)$ has finite value.}
\label{fig:logistic_function}
\end{figure}

Our goal is to online estimate whether one or more motors have failed (consequently limiting maximum thrust and moments). To do so, we introduce the health variable $h_i \in \mathbb{R}$ for each individual motor $i$ and assume that the effective force generated from the $i^{\text{th}}$ motor is $f_i^e = L(h_i)f_i$, where $L(h) = \frac{1.05}{1+e^{-h}}$ is the logistic function shown in Figure \ref{fig:logistic_function} and $f_i$ corresponds to the respective motor thrust. Intuitively, we expect that $L(h_i) = 1$ for a healthy motor and $L(h_i)\rightarrow0$ for a stopped one. 
We implemented an \ac{EKF} that estimates the set of health variables $h_i$ online (and thus the motor thrust $f_i$) for each individual motor. 
The effective body torques and collective thrust are now given by:
\begin{equation}
    \begin{bmatrix}
     \myvec{M}{B} \\ 
     T 
    \end{bmatrix} = \mathbf{A}  \begin{bmatrix} 
                                    L(h_1) f_{1} & \cdots & L(h_6) f_{6} 
                                    \end{bmatrix}^{\top},
\end{equation}
where $\mathbf{A}$ is the allocation matrix defined in \eqref{eq:allocation_matrix}, which depends on the moment coefficients and the motor direction of rotation.
% Logistic function and Health variables derivative

In our \ac{EKF}, we use the following prediction and observation models:
\begin{subequations}
\begin{align}
     \rotveldot{B}{}{} &= \mathbf{J}^{-1} (\myvec{M}{B} - \rotvel{B}{}{} \times \mathbf{J}\rotvel{B}{}{}) + \bm{w}_{\omega},  \\
     \dot{h}_i &= \frac{1}{\tau_h}(\Bar{h} - h_i)+ w_h, \\
     \dot{f}_i &= \frac{1}{\tau_f}(f_i^r - f_i + w_f), \\
     \myvec{z}{}:&= \begin{bmatrix}
                \rotvel{B}{}{}  + \bm{v}_{\tilde{\omega}} \\
                T  + v_{T}
              \end{bmatrix}.
\end{align}
\end{subequations}
The noises $\bm{w}_{\omega}$,  $w_h$, and $w_f$ are Gaussian white noise processes with densities $\sigma_\omega$, $\sigma_h$, and $\sigma_f$, respectively; the measurement $\mbf{z}$ is assumed to be corrupted by $\bm{v}_{\tilde{\omega}} \sim \mathcal{N}(0,\,\sigma^{2}_{\tilde{\omega}}\mathbf{I})$ and $v_{T}\sim \mathcal{N}(0,\,\sigma^{2}_{\tilde{T}})$. Furthermore, $f_i^r$ stands for the per-motor reference thrust as given by the control allocation and $\tau_f$ for the time constant characterising the first order motor thrust dynamics.  The measurements $\rotveltilde{B}{}{}$ and  $\tilde{T}$ required for the \ac{EKF} update are obtained using the onboard \ac{IMU}. For $\rotveltilde{B}{}{}$ we use the bias corrected gyro measurements and for measured collective thrust $\tilde{T}$ we use $\tilde{T}\approx m a_z$ with $m$ denoting the known mass of the \ac{MAV} and $a_z$ the accelerometer measurement along the z axis. Notice that our observation model for the collective thrust does not account for the  $\rotvel{B}{}{}\times\vel{B}{}{}$ term which appears in the Body frame expressed linear acceleration dynamics. The values for the noise parameters and model constants are given in Table \ref{table:EKF_parameters}.
\begin{table}[ht]
\caption{\ac{EKF} Parameters}
\centering
\resizebox{\columnwidth}{!}{%
\begin{tabular}{l l l l l l l l}
$\sigma_{\omega}$ & $3.16~\mathrm{rad}/(\mathrm{s}\sqrt{\mathrm{hz}})$ & $\sigma_h$  
& $0.31~\sqrt{\mathrm{hz}}^{-1}$  & $\sigma_f$  & $0.94~\mathrm{N}/\sqrt{\mathrm{hz}}$ 
& $\sigma_{\tilde{\omega}}$ & 0.01 $\mathrm{rad}/\mathrm{s}$ \\
$\sigma_{\tilde{T}}$ & $0.1$ N & $\tau_h$ & $0.3~\si{\second}$ & $\tau_f$ & $0.01~\si{\second}$ & $\bar{h}$ &  $2.99$
\end{tabular}
}
\label{table:EKF_parameters} 
\end{table}

In order to avoid false positives due to e.g. inaccurate model, we use the estimated value of $h_i$ and its estimated uncertainty. We thus consider a motor failed when $L(h_i +3\sigma_i)<0.5$ (with $\sigma_i$ denoting the health state standard deviation obtained as a marginal from the state covariance matrix). When the above inequality is true we update the control allocation algorithm by setting $f_{\text{min}} = f_{\text{max}}= 0$ for the failed motor and enabling the bidirectional mode for the opposite.
\section{Experiments}
We showcase the capabilities of our algorithms in two different scenarios, namely response to step commands and autonomous detection and recovery after a motor failure.
For the experiments presented we used a custom-built hexacopter using off-the-shelf components. It consists of a 550 mm wide frame, a Pixhawk flight controller flashed with a modified version of the PX4 firmware and an Intel NUC-7567U onboard computer. We used 960KV motors coupled with the carbon reinforced Aeronaut CAMcarbon $9.5\times4.5$ propellers and the DYS ARIA bidirectional capable ESCs. A Vicon motion capture system was responsible for providing external position and orientation measurements while all the other components run onboard the \ac{MAV}.
\subsection{Aggressive step commands}
\begin{figure}[htbp]
\includegraphics[width=0.228\textwidth]{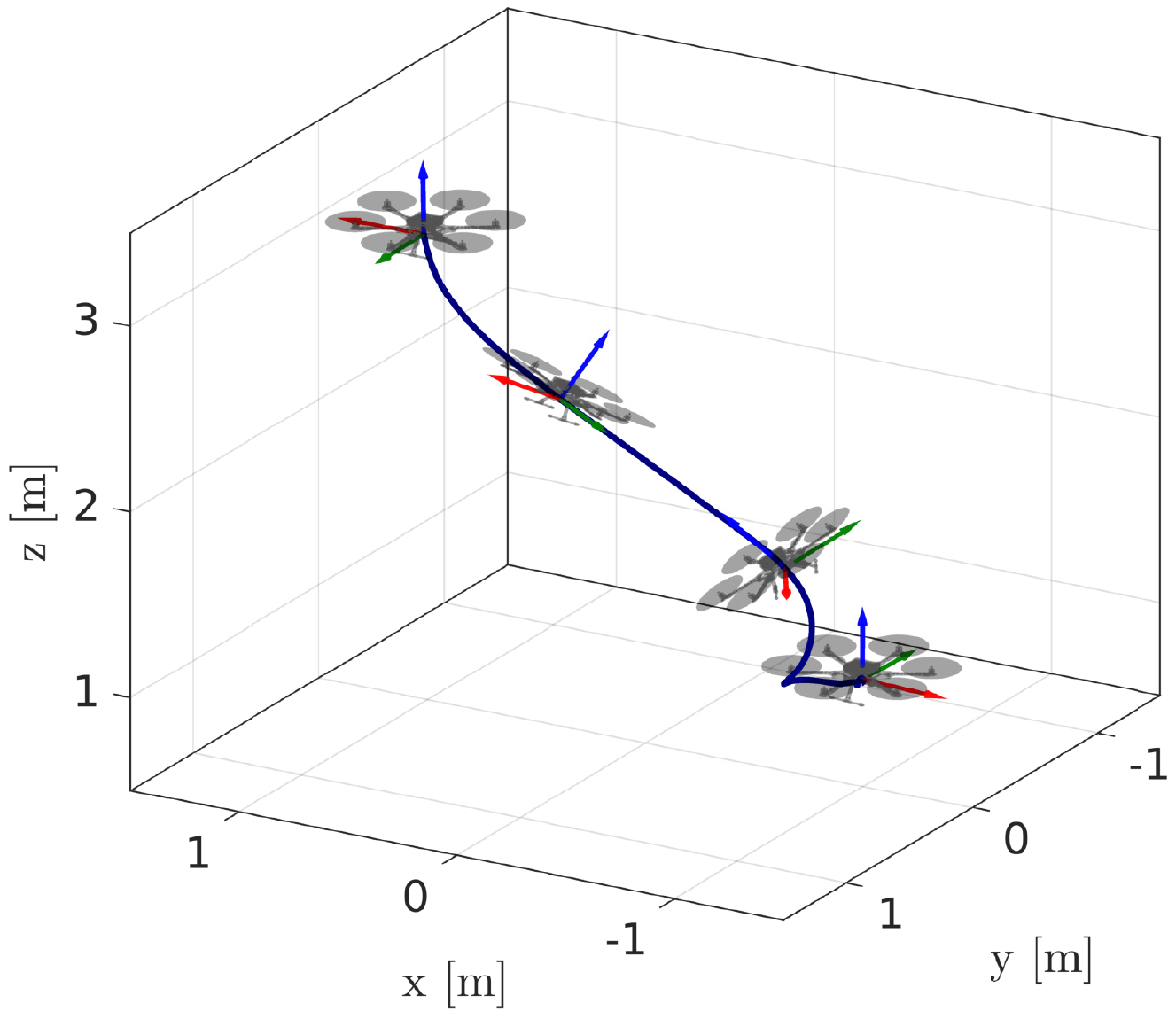}
\includegraphics[width=0.24\textwidth]{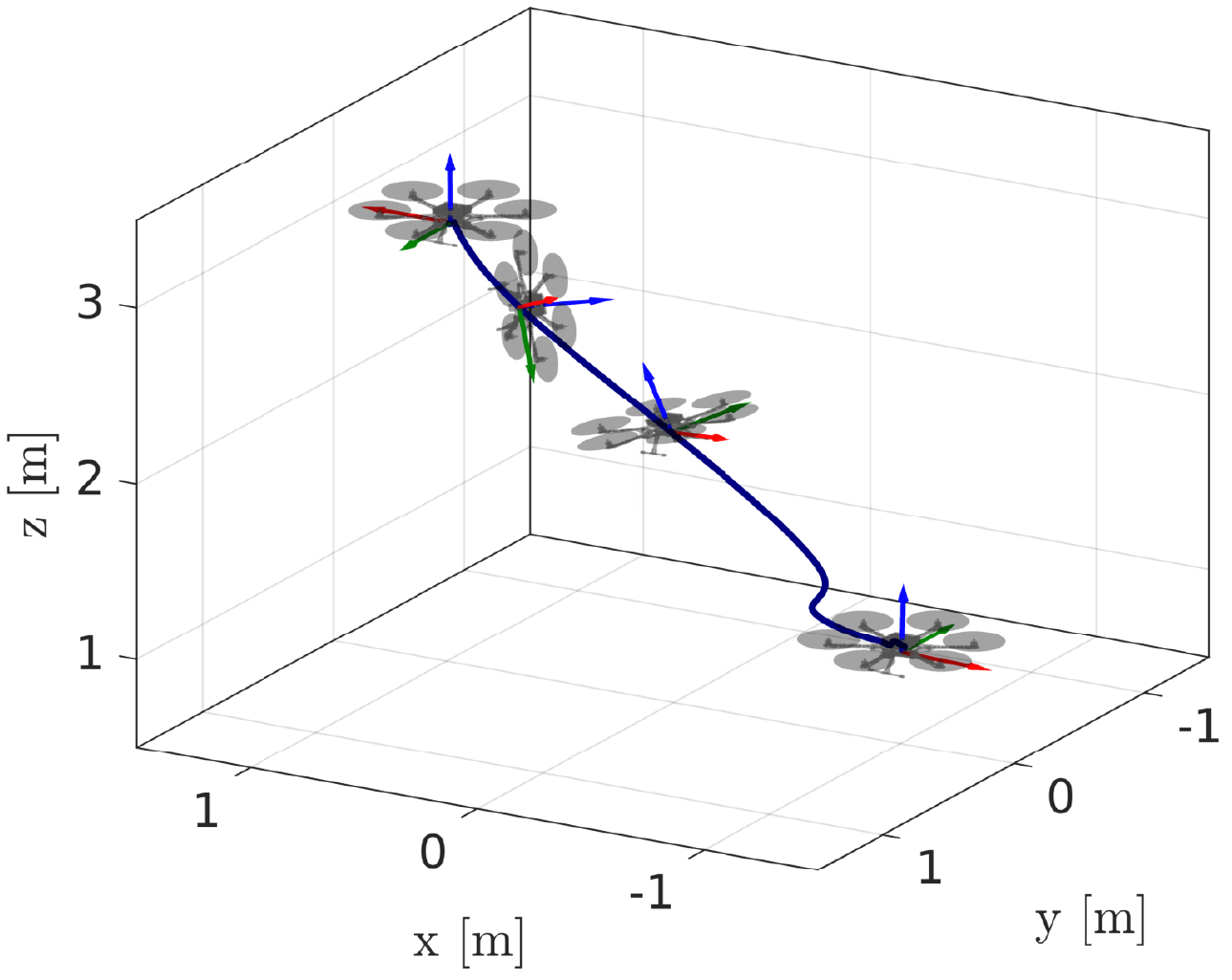}
\caption{Our \ac{MAV} executing a 2 \si{\meter}~step in x and z and a 180$^o$ step in yaw with low (left) and high (right) orientation gains. Peak acceleration exceeds 15$\si{\meter}/\si{\second}^2$ while peak orientation exceeds 120$^o$ in pitch.}
\label{fig:position_step}
\end{figure}

In order to verify the tracking capabilities of the designed \ac{NMPC}, Figure \ref{fig:position_step} shows the \ac{MAV} response for a 2 \si\meter~step in x and z and a 180$^o$ step in yaw. The \ac{NMPC} generated dynamically feasible trajectories which can steer the \ac{MAV} in any orientation and achieve large linear accelerations (given the physical limitations of the platform) without overshooting. We conducted the same experiment twice using low and high orientation gains and observed that, in the latter case, the \ac{NMPC} reduces the yaw error faster by simultaneously performing a half flip in roll and pitch. 
\subsection{Fault detection and recovery}
\begin{figure*}[htbp]
\centering
\includegraphics[width=0.31\textwidth]{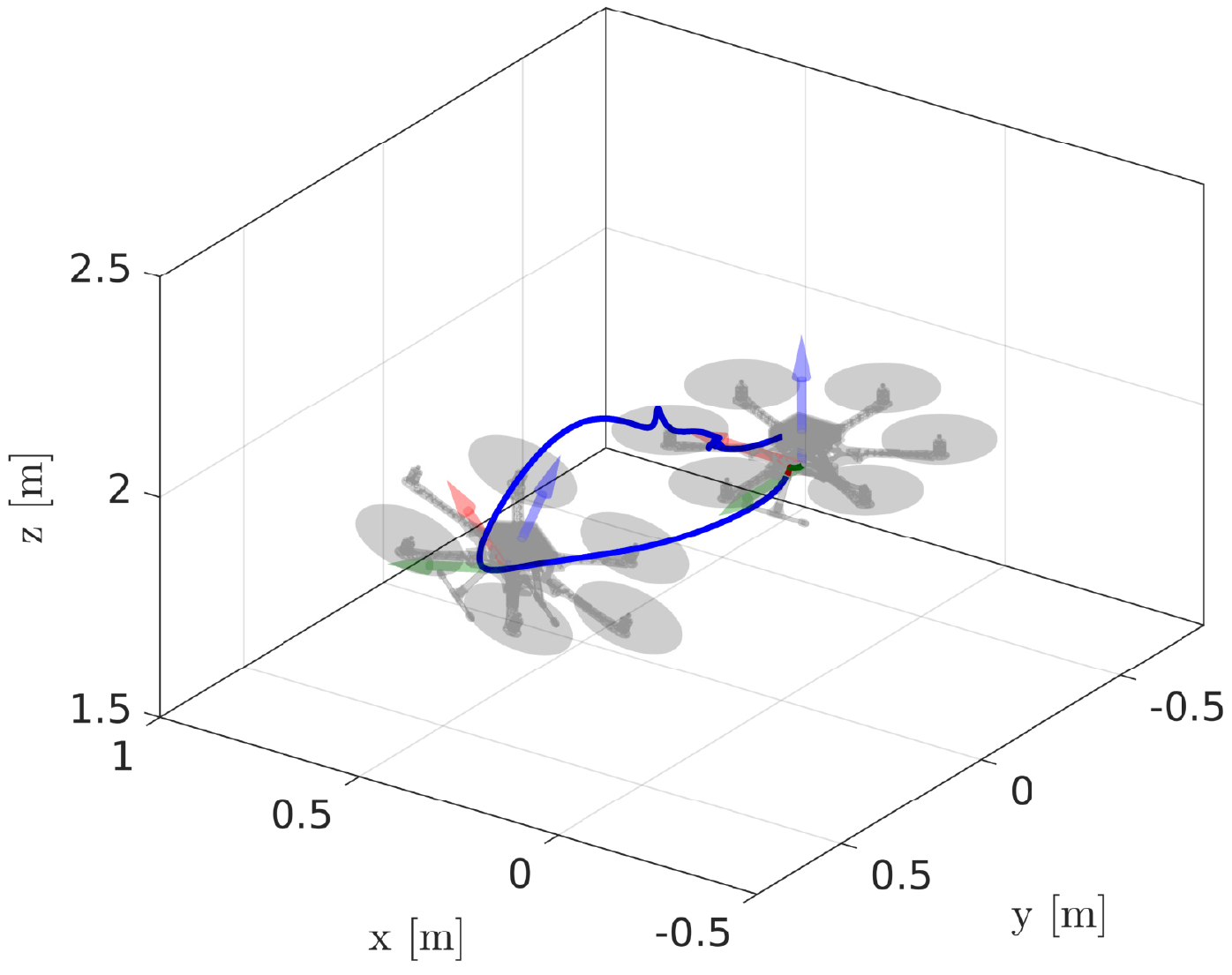}
\includegraphics[width=0.31\textwidth]{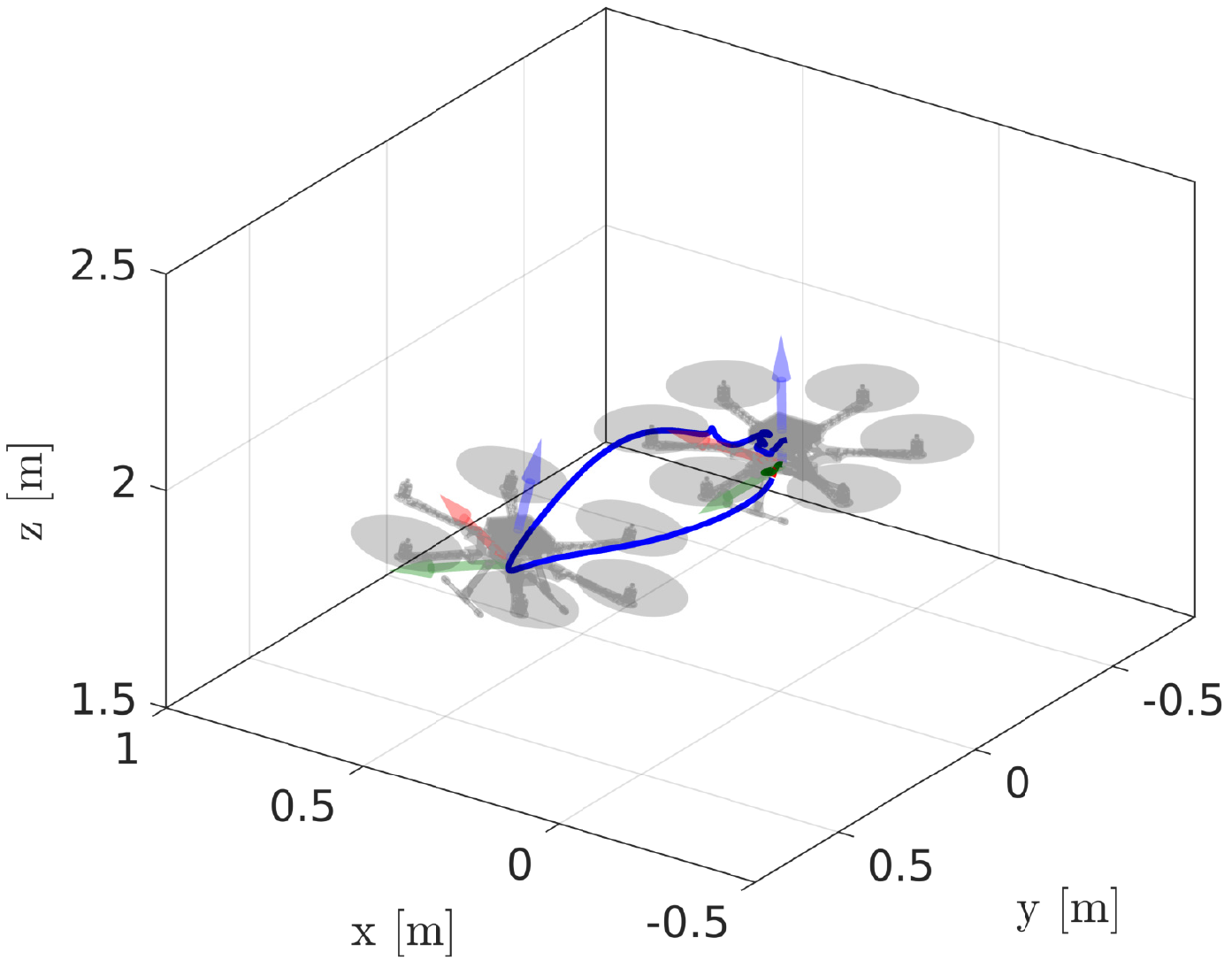}
\includegraphics[width=0.32\textwidth]{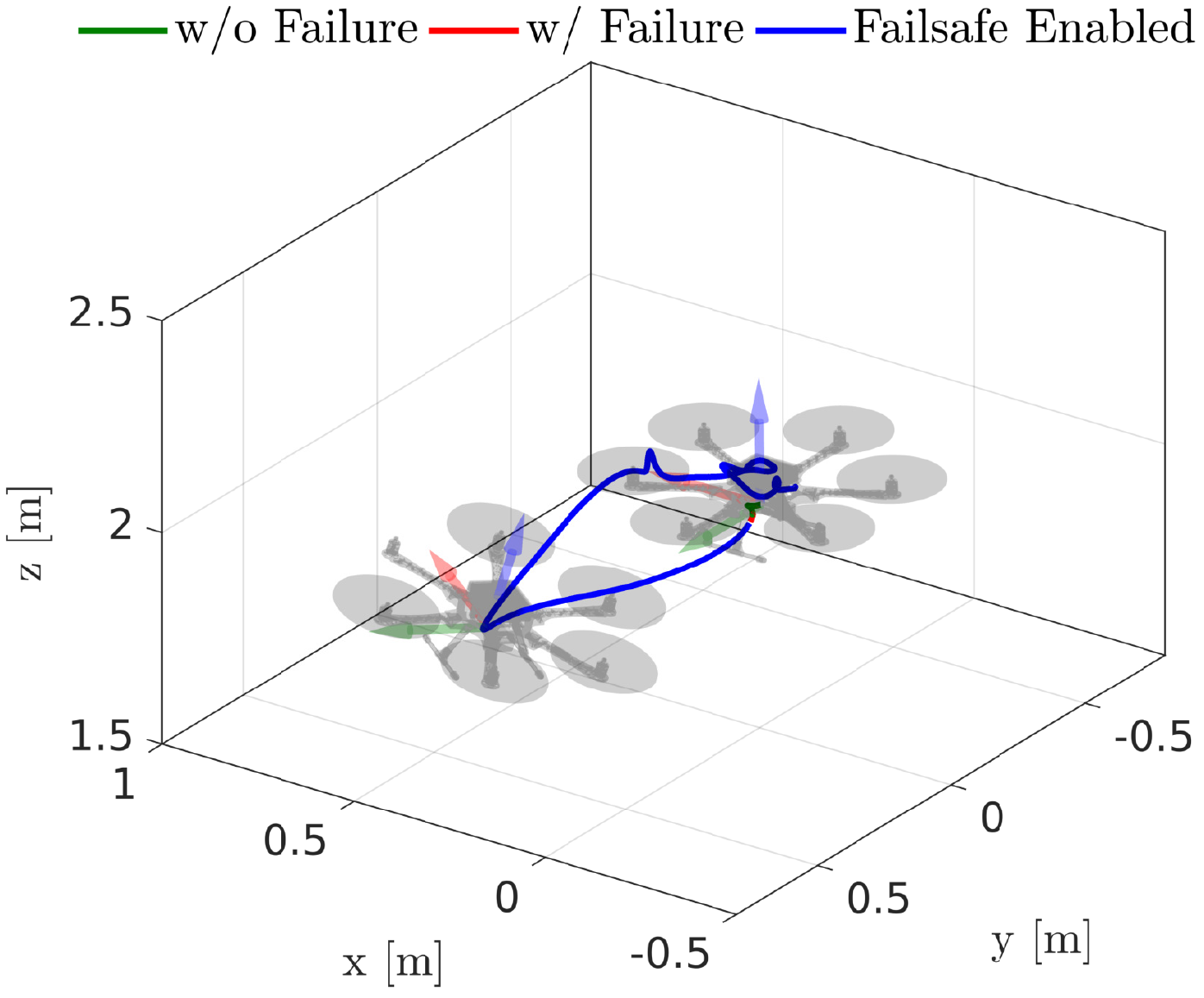}\\
\includegraphics[width=0.315\textwidth]{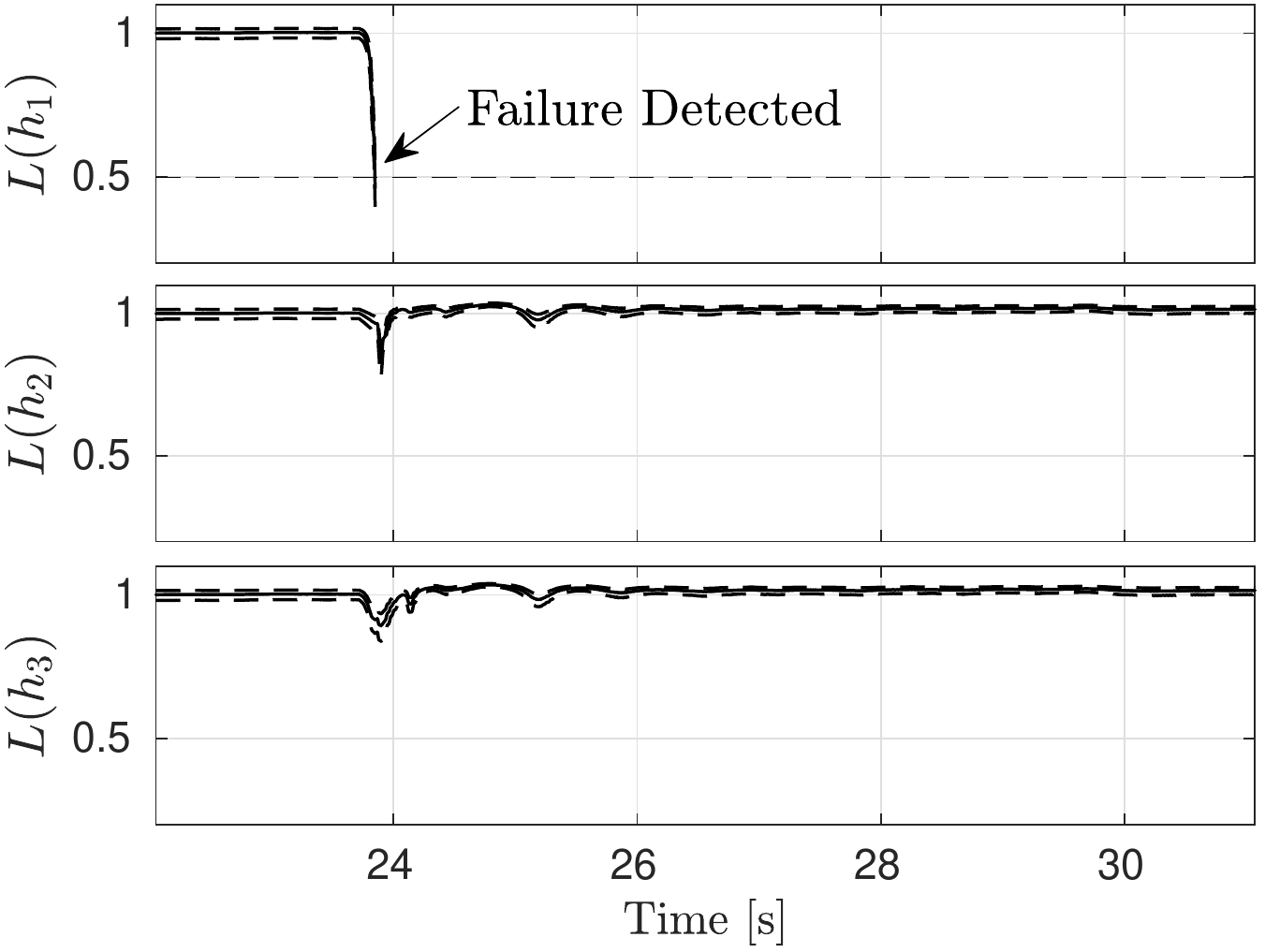}
\includegraphics[width=0.32\textwidth]{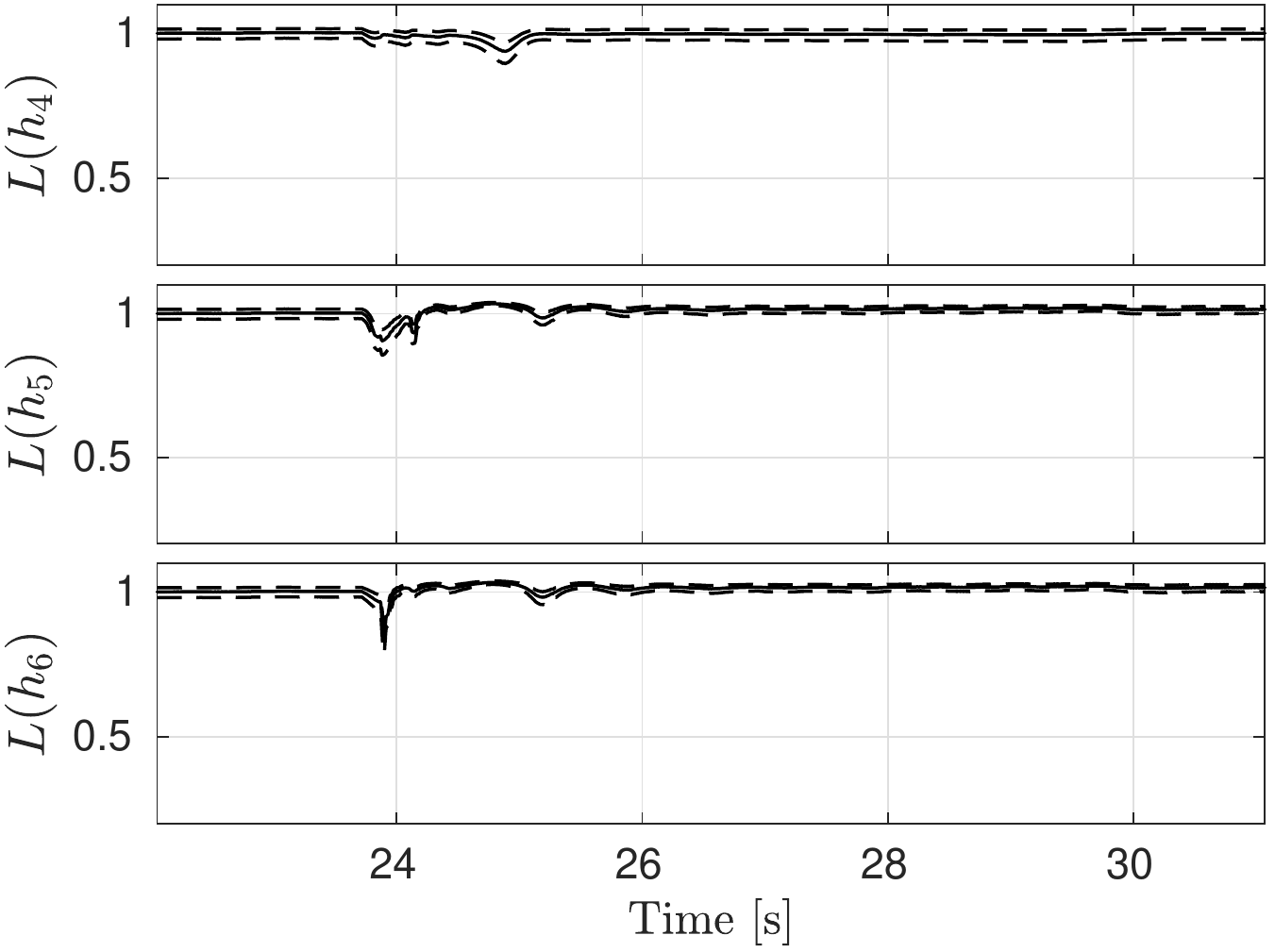}
\includegraphics[width=0.28\textwidth]{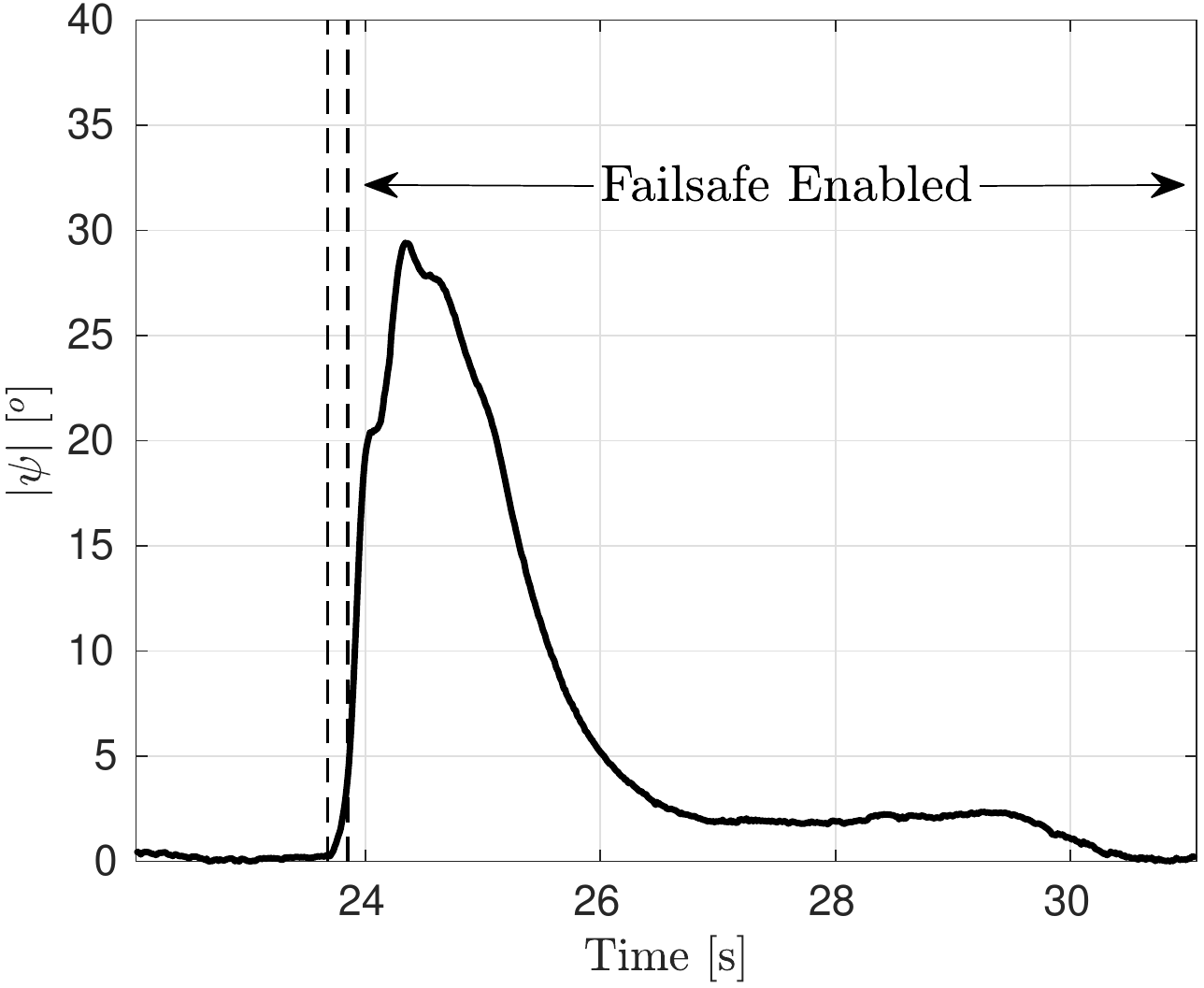}
\caption{Top row: Three different experiments with autonomous fault identification and recovery during hover. In all the experiments the failure was identified and the fail-safe was triggered within 0.18 \si{\second} after the manual deactivation of Motor 1. The \ac{MAV} was able to recover with a maximum height loss of 0.40 \si{\meter}. Bottom row: The online estimates of the health status $L(h_i)$ and their corresponding $3\sigma$ confidence bounds and the absolute yaw error for the first experiment. Notice how the upper bound estimate $L(h_1+3\sigma_1)$ for Motor 1 drops below the 0.5 threshold after the motor deactivation at t = 23.68\si{\second}. Once the fail-safe is triggered at t = 23.85\si{\second}, control of yaw (bottom right) is maintained and the error converges to zero.}
\label{fig:hover_failsafe}
\end{figure*}

\begin{figure*}[htbp]
\centering
\includegraphics[width=0.30\textwidth]{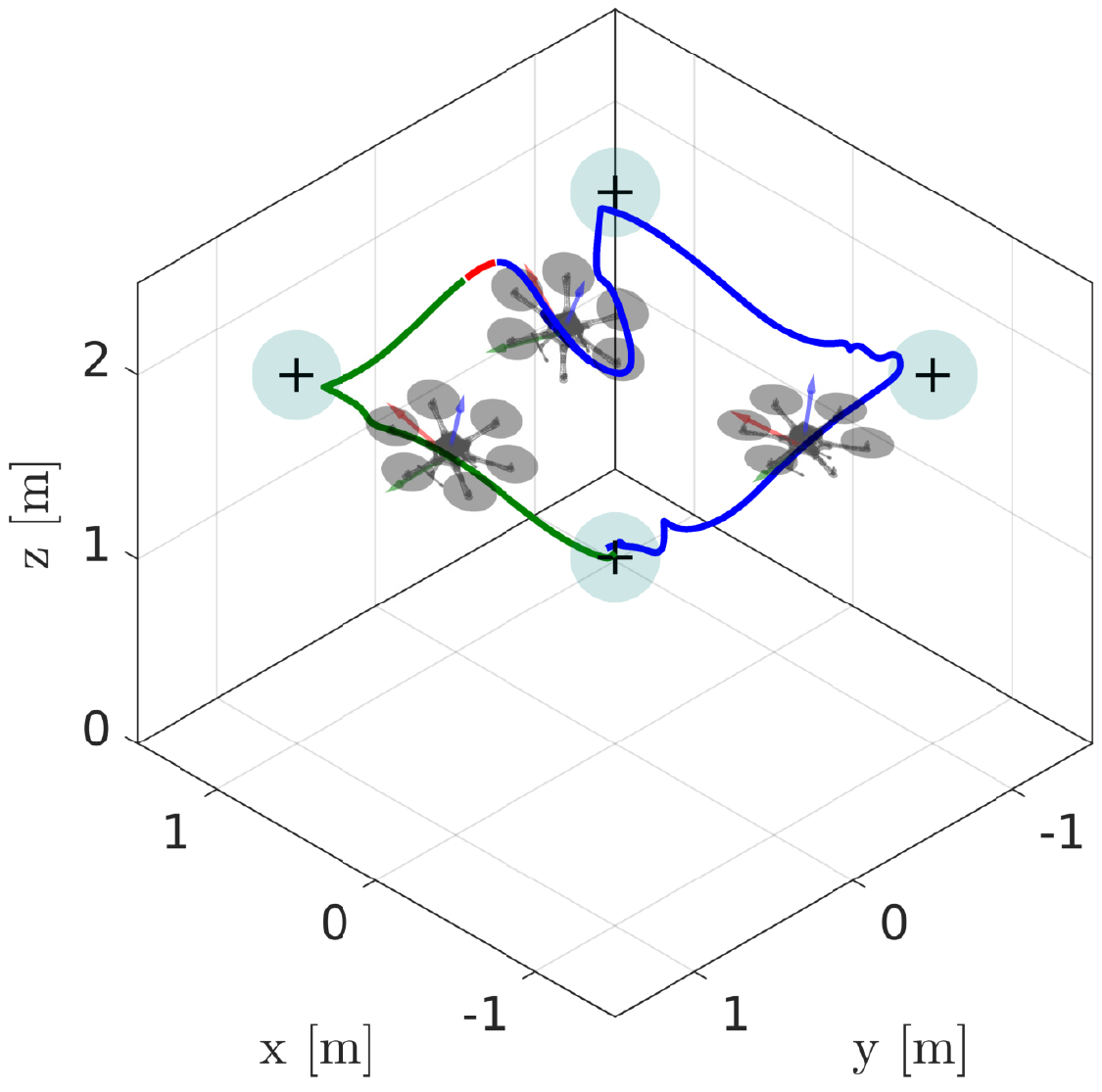}
\includegraphics[width=0.30\textwidth]{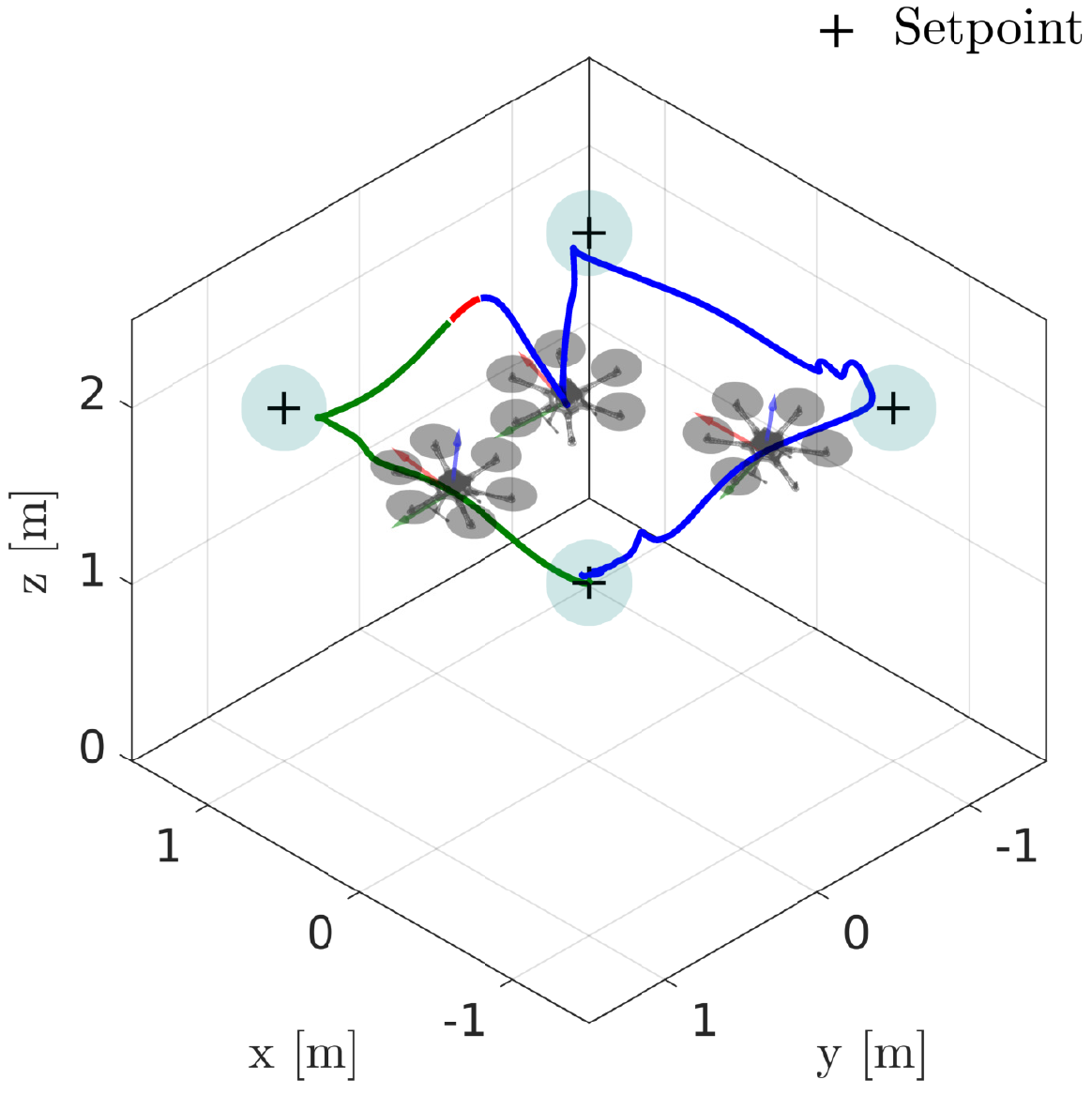}
\includegraphics[width=0.34\textwidth]{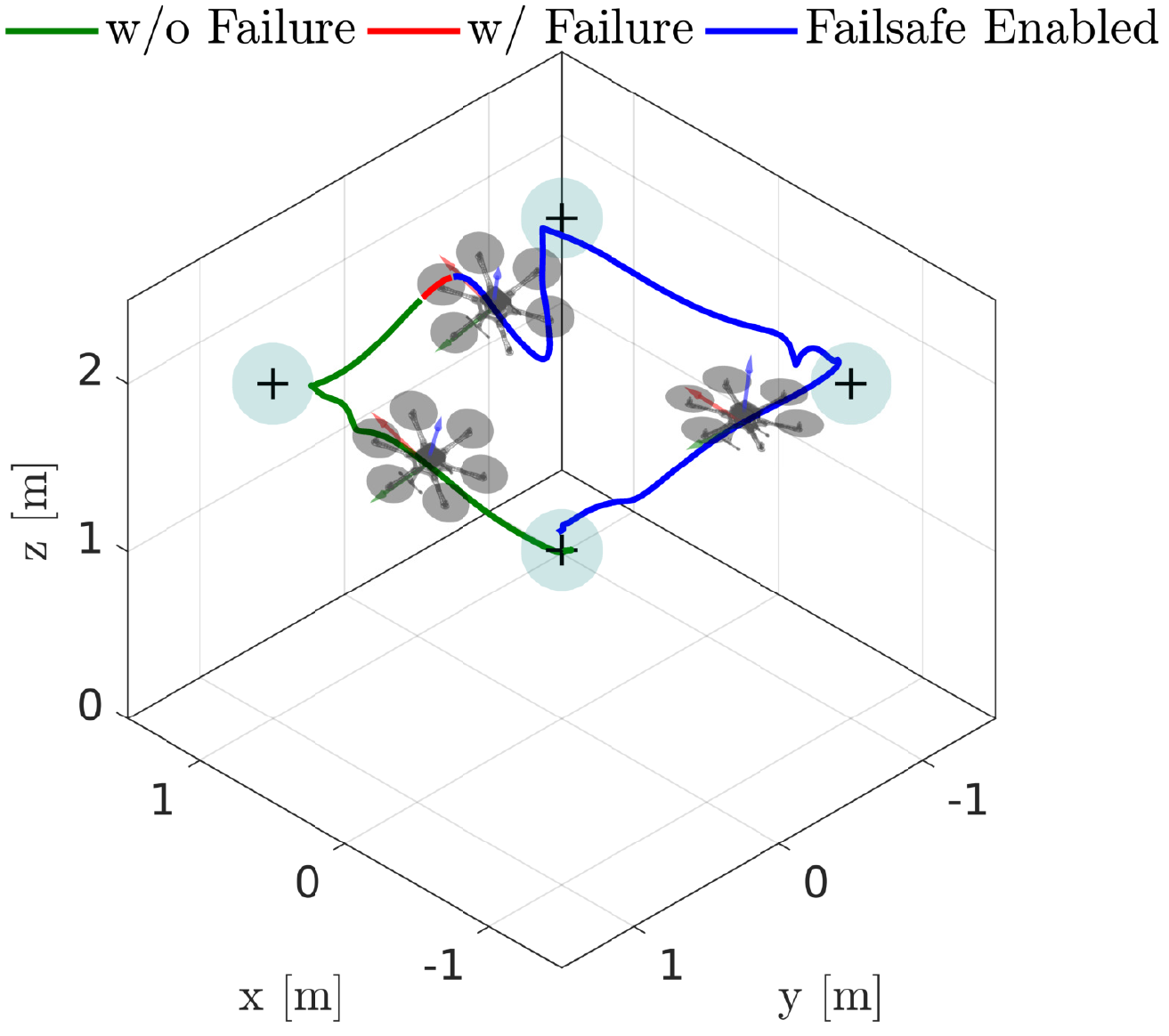}\\
\includegraphics[width=0.33\textwidth]{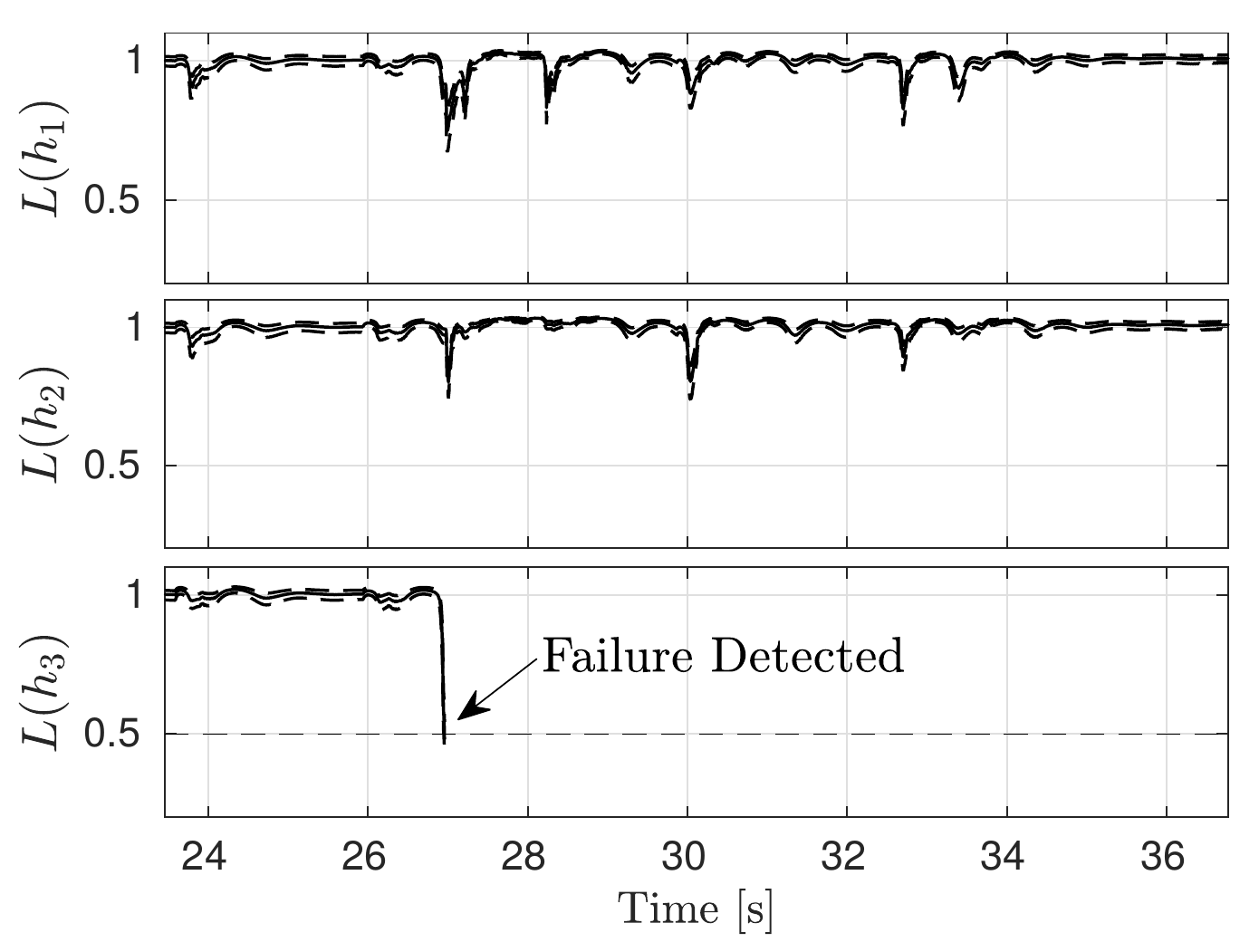}
\includegraphics[width=0.32\textwidth]{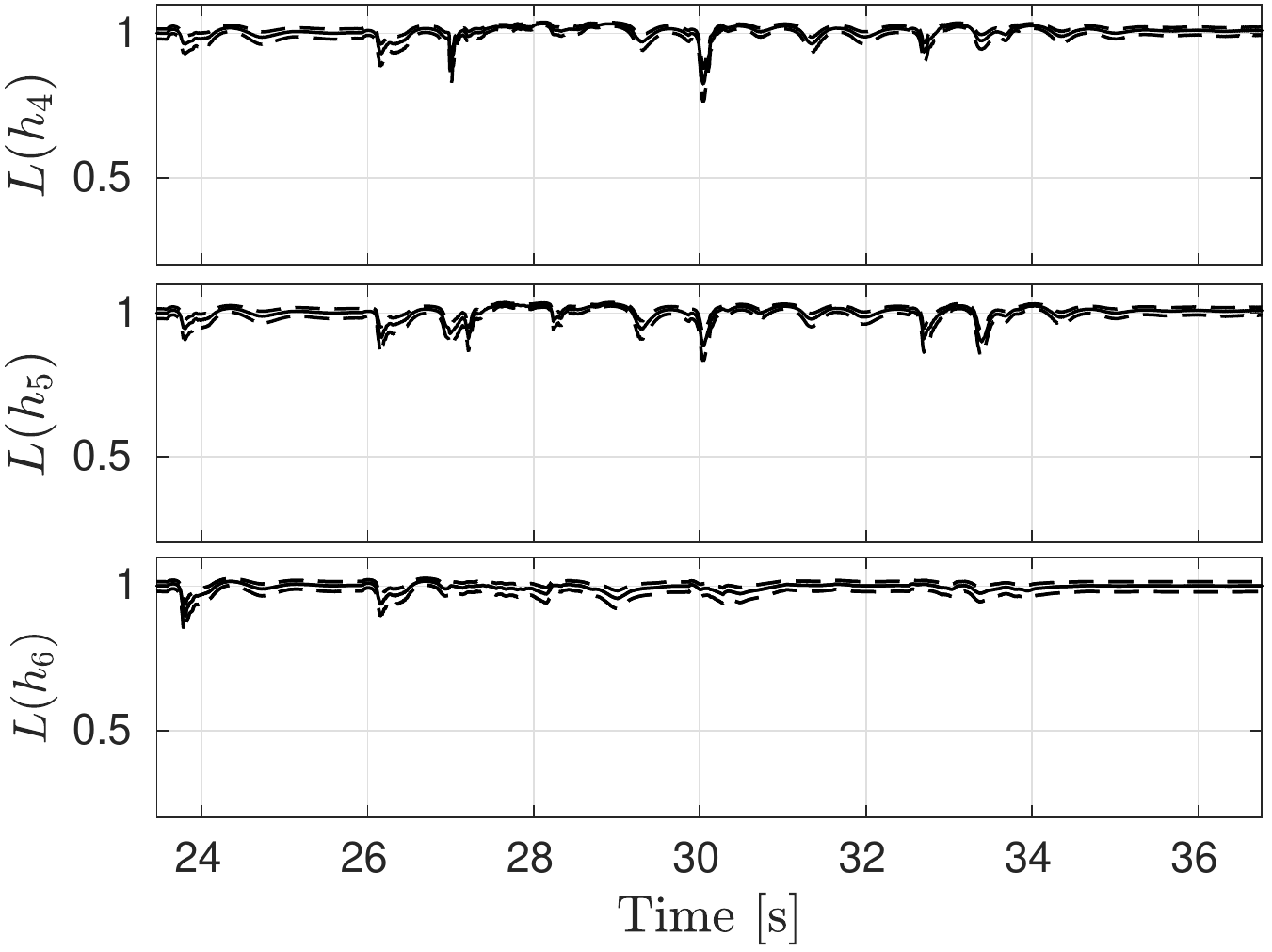}
\includegraphics[width=0.27\textwidth]{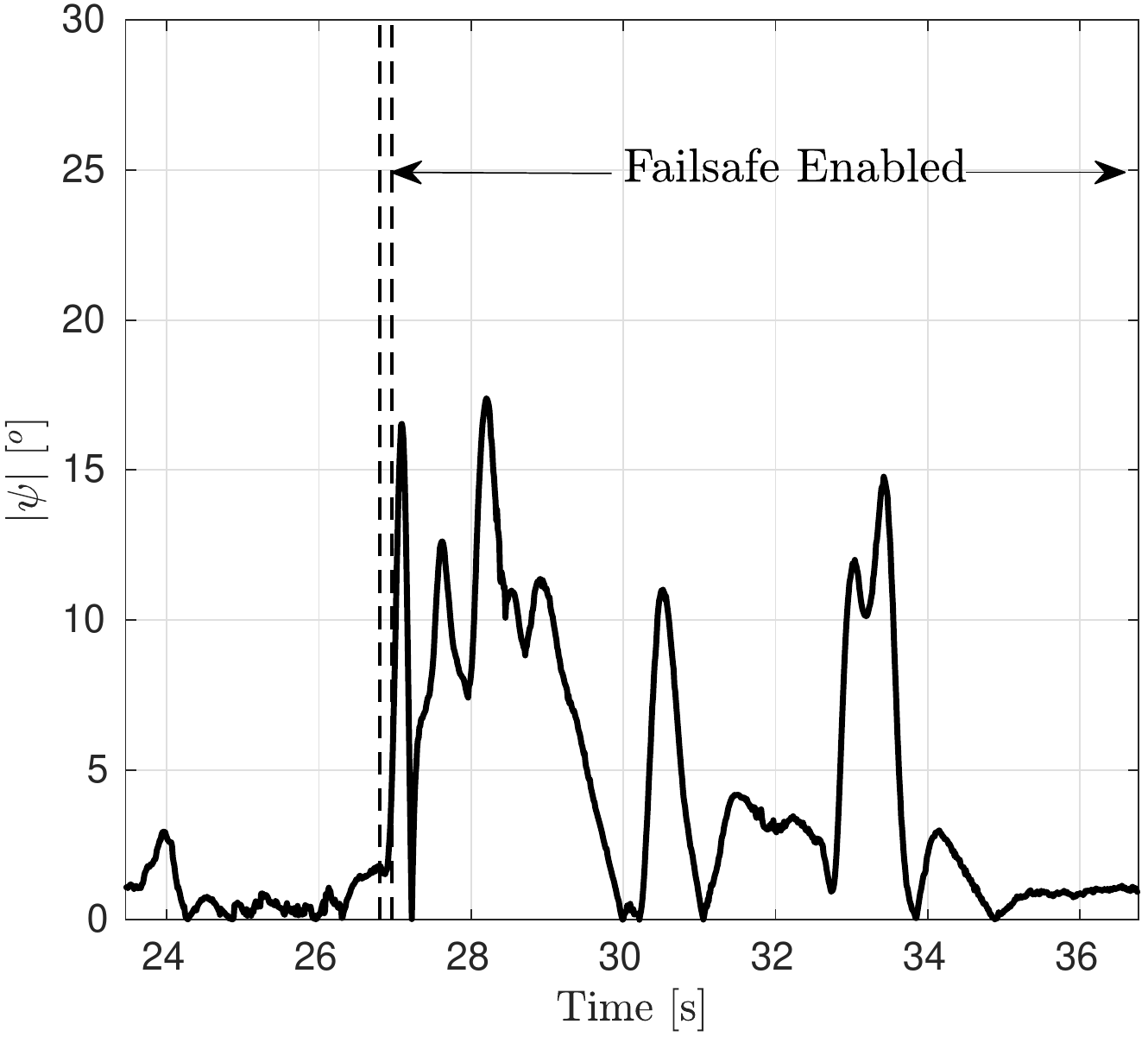}\\
\caption{Top row: Three different experiments with autonomous fault identification and recovery while following setpoint commands. In all the experiments the failure was identified and the fail-safe was triggered within 0.18 \si{\second} after the manual deactivation of Motor 3. The \ac{MAV} was able to recover with a maximum height loss of 0.60 \si{\meter}. Bottom row: The online estimates of the health status $L(h_i)$ and their corresponding $3\sigma$ confidence bounds and the absolute yaw error for the first experiment. The upper bound estimate $L(h_1+3\sigma_1)$ for Motor 3 drops below the 0.5 threshold after the motor deactivation at t = 26.80 \si{\second}. Once the fail-safe is triggered at t = 26.96 \si{\second}, control of yaw (bottom right) is maintained and the error converges to zero.}
\label{fig:square_failsafe}
\end{figure*}

We tested the failure detection and autonomous recovery in two different scenarios where one motor was switched off (i) while hovering and another (ii) while the \ac{MAV} was following setpoint commands. The results regarding the position and yaw tracking along with the online estimated health status of each motor, are shown in Figures \ref{fig:hover_failsafe} and \ref{fig:square_failsafe}. The injected motor failure was correctly identified with a maximum delay of $0.18~\si{\second}$. In both scenarios, the  \ac{MAV} was able to recover with a maximum height loss of $0.6~\si{\meter}$. Position and yaw references were still tractable however the 5-motor asymmetric configuration resulted in slower tracking response. Regarding the health status variables of the functioning motors, these always remain close to 1. It can be seen that, in the setpoint experiment, there exist some short-in-duration deviations from 1. These spikes correspond to time instants when large angular accelerations were executed.  We consider the main reason for this behaviour to be the mismatch between the \ac{EKF} prediction model (which does not take into account less significant phenomena, such as gyroscopic moments) and the real one. In any case the estimated upper bound was always greater than 0.8 and thus unable to trigger a false positive.
\section{Conclusion and Future Work}
This paper presented a series of algorithms that can be used for aggressive and fault tolerant multicopter navigation. We experimentally verified their performance using a hexacopter although the same approach can be implemented on any other multirotor with minor modifications. % such as a quadrotor with minor modifications(e.g.\ fly with free heading in the event of a motor failure).  
Control performance can be further improved by using a more accurate system model, as the current one does not take into account effects such as the motor dynamics, rotor drag and gyroscopic moments. The fault detection \ac{EKF} can seamlessly be implemented as an algorithmic update on any \ac{MAV} as it only requires inertial measurements. It can be accordingly extended with speed or current measurements in order to prevent false positives due to large external disturbances. The disadvantage of our approach is that it requires carefully identifying physical parameters such as the motor coefficients and the inertia tensor. By online estimating these as shown in \cite{Burri:JFR2018}, we can make the controller adaptive to model changes and eliminate the need for tedious accurate offline identification. 
%\addtolength{\textheight}{-12cm}   % This command serves to balance the column lengths
                                  % on the last page of the document manually. It shortens
                                  % the textheight of the last page by a suitable amount.
                                  % This command does not take effect until the next page
                                  % so it should come on the page before the last. Make
                                  % sure that you do not shorten the textheight too much.

%%%%%%%%%%%%%%%%%%%%%%%%%%%%%%%%%%%%%%%%%%%%%%%%%%%%%%%%%%%%%%%%%%%%%%%%%%%%%%%%

%%%%%%%%%%%%%%%%%%%%%%%%%%%%%%%%%%%%%%%%%%%%%%%%%%%%%%%%%%%%%%%%%%%%%%%%%%%%%%%%

%%%%%%%%%%%%%%%%%%%%%%%%%%%%%%%%%%%%%%%%%%%%%%%%%%%%%%%%%%%%%%%%%%%%%%%%%%%%%%%%
%\section*{Appendix}

%\section*{Acknowledgements}

\bibliographystyle{IEEEtran}
\bibliography{./bibliography/main.bib}

\end{document}